\documentclass[final,3p,times,twocolumn]{elsarticle}

\usepackage{booktabs}
\usepackage{hyperref}
\usepackage{amssymb}
\usepackage{amsmath}
\usepackage{array}
\usepackage{xcolor}
\usepackage{lineno}
\usepackage{graphicx} 
\usepackage{booktabs} 
\usepackage{array}    
\usepackage{caption}  

\newcommand{\kl}[1]{#1}
\newcommand{\klqs}[1]{}

\begin{document}

\begin{frontmatter}



\title{Denoising-While-Completing Network (DWCNet): Robust Point Cloud Completion Under Corruption}

\author[iinst1]{Keneni W. Tesema\corref{cor1}}
\ead{k.w.tesema@swansea.ac.uk}
\cortext[cor1]{Corresponding author, ORCID: \url{https://orcid.org/0009-0003-1247-2435}}


\author[iinst2]{Lyndon Hill}
\ead{l.hill@ieee.org}

\author[iinst1]{Mark W. Jones}
\ead{m.w.jones@swansea.ac.uk}

\author[iinst1]{Gary K. L. Tam}
\ead{k.l.tam@swansea.ac.uk}

\affiliation[iinst1]{organization={Swansea University},
           addressline={Fabian Way},
             city={Swansea},
            postcode={SA1 8EN},
            state={Wales},
             country={United Kingdom}}

\affiliation[iinst2]{organization={Beam},
            addressline={Bristol},
            state={England},
            country={United Kingdom}}


\vspace{-0.1 in}
\begin{abstract}

Point cloud completion is crucial for 3D computer vision tasks in autonomous driving, augmented reality, and robotics. However, obtaining clean and complete point clouds from real-world environments is challenging due to noise and occlusions. Consequently, most existing completion networks -- trained on synthetic data -- struggle with real-world degradations. In this work, we tackle the problem of completing and denoising highly corrupted partial point clouds affected by multiple simultaneous degradations. To benchmark robustness, we introduce the Corrupted Point Cloud Completion Dataset (CPCCD), which highlights the limitations of current methods under diverse corruptions. Building on these insights, we propose DWCNet (Denoising-While-Completing Network), a completion framework enhanced with a Noise Management Module (NMM) that leverages contrastive learning and self-attention to suppress noise and model structural relationships. DWCNet achieves state-of-the-art performance on both clean and corrupted, synthetic and real-world datasets. The dataset and code will be publicly available at \url{https://github.com/keneniwt/DWCNET-Robust-Point-Cloud-Completion-against-Corruptions}. 

\end{abstract}







\end{frontmatter}




\section{Introduction}
\label{sec1:intro}

Point cloud completion is the task of completing a partial point cloud input so that it fully represents a 3D shape \cite{dldl, mytvcg}. It is a critical component in various tasks such as object recognition, 3D reconstruction, and point cloud pre-training \cite{dldl, mytvcg}. These tasks have practical applications in diverse fields including autonomous driving \cite{auto, auvreg}, robotics \cite{grasp}, and augmented reality \cite{arch}, among others.

\begin{figure}[th!]
\centering
\includegraphics[width=\linewidth]{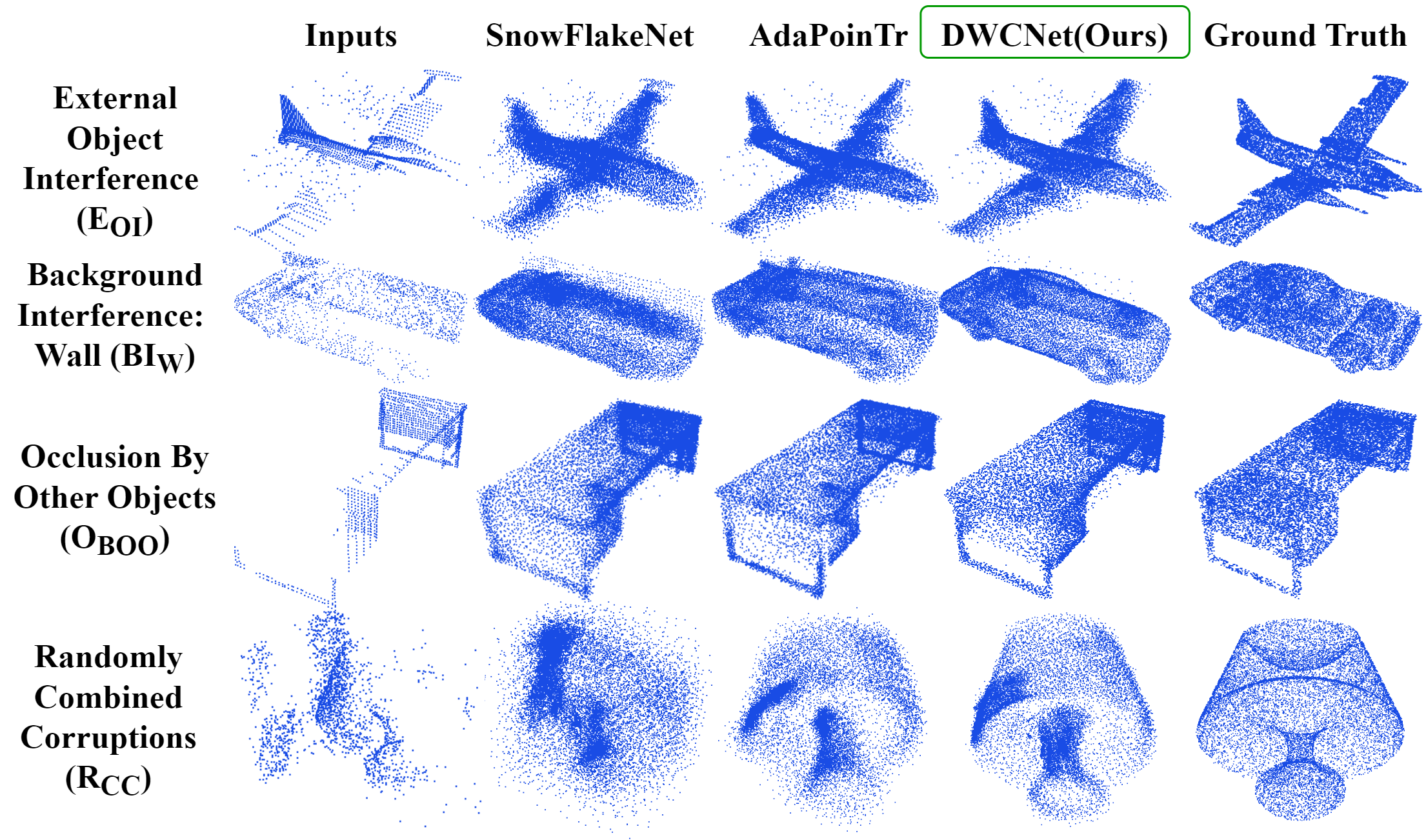} 
\vspace{-0.15in}
\caption{\label{teaserpic}We propose a corrupted point cloud completion benchmark dataset (CPCCD) and a robust point cloud completion network, DWCNet.}
\vspace{-0.25 in}
\end{figure}

In recent years, research in point cloud completion has witnessed considerable advancements, with a primary focus on deep learning-based techniques \cite{dldl, mytvcg}. Algorithms like ODGNet \cite{odgnet} and AdaPoinTr \cite{adapointr} have demonstrated promising results on datasets such as PCN \cite{pcn}, MVP \cite{vrcnet} and KITTI \cite{kitti}. These techniques heavily depend on comprehensive training data from datasets like ShapeNet \cite{shapenet}.

Supervised point cloud completion differs from other point cloud learning tasks such as segmentation or object detection in the type of supervision it requires. While segmentation and detection rely on labeled annotations (e.g., class labels or bounding boxes), completion networks require clean and complete 3D point clouds as ground truth during training. However, obtaining such high-quality reference data from real-world scans is extremely challenging. As a result, most existing methods are trained and evaluated on synthetic datasets. Although these datasets are diverse and consistent \cite{pcn, topnet, vrcnet}, they are typically generated by uniformly sampling 3D meshes \cite{trainingdatasurvey}, which makes them overly clean and unrepresentative of the noise and artifacts found in real-world data \cite{pointcloudc}. This leads to two key observations: (1) the lack of a benchmark dataset designed to systematically evaluate robustness against corrupted inputs, and (2) the inability of current completion methods to generalize well to real-world-like corruptions.

There have been limited attempts to incorporate noise and corruptions into partial point clouds for completion \cite{multimodal, adapointr}. 
However, these efforts address only Gaussian noise, which represents just one of the many potential corruptions in real-world point clouds. Furthermore, the fidelity of synthetic partial point clouds used to train completion algorithms remains subpar compared to real-world scans. Even the real-world dataset most commonly used for evaluation, KITTI \cite{kitti}, lacks environmental corruptions \cite{robo3d}. Point clouds from the ScanNet dataset \cite{scannet}, often used for evaluation \cite{p2c, pointattn}, are preprocessed before completion, and the results are typically qualitative, leading to subjective assessments. Thus, there is a clear need for a corrupted benchmark dataset to enable objective and qualitative robustness evaluation.

\begin{figure}[t!]
\centerline{\includegraphics[width=\linewidth]{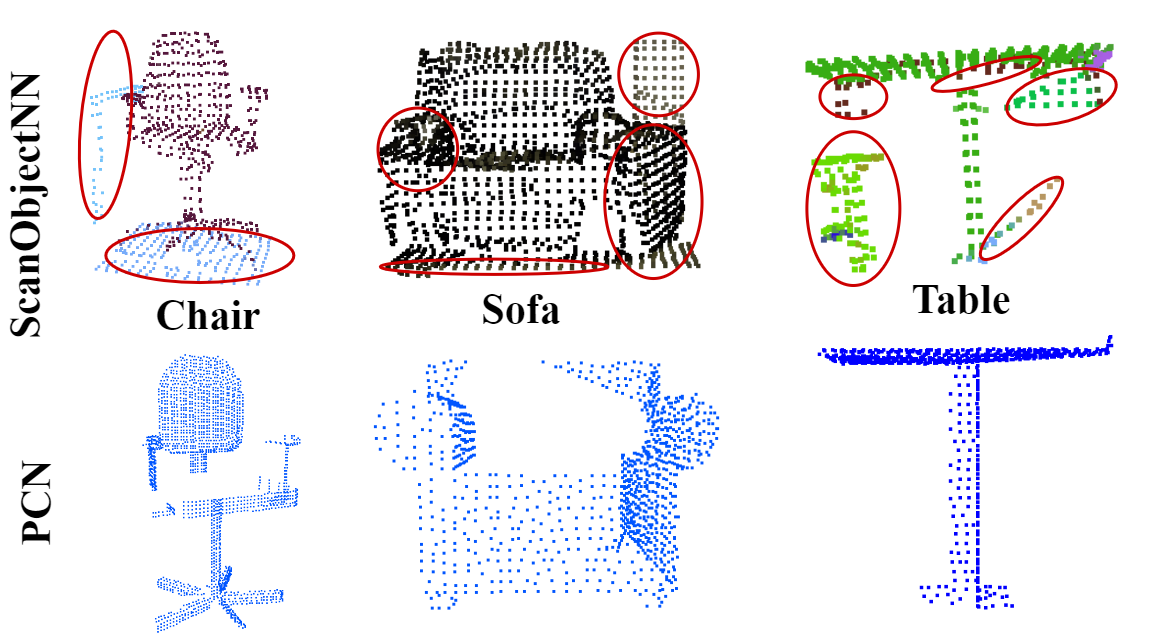}}
\caption{Real world scans (like ScanObjectNN dataset) often contain corruptions (shown in the ellipses) while point clouds in point cloud completion datasets (like PCN) are often clean.}
\label{scanobj}
\vspace{-0.18 in}
\end{figure}

\begin{figure*}[th!]
\vspace{-0.1in}
\centerline{\includegraphics[width=\linewidth]{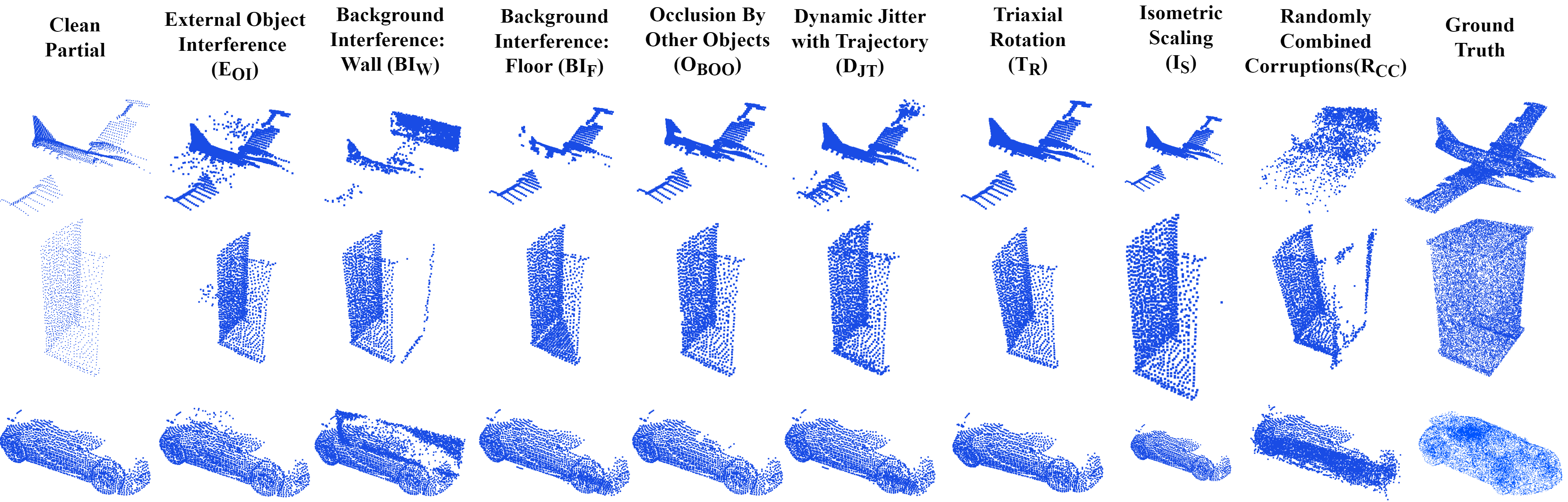}}
\caption{CPCCD Dataset and its different types of corruptions} 
\label{cpccdfig}
\vspace{-0.1in}
\end{figure*}

To address the first issue, we introduce the Corrupted Point Cloud Completion Dataset (CPCCD) (Figure~\ref{teaserpic}). We classify the corruptions observed in the real-world ScanObjectNN dataset \cite{scanobjectnn} (Figure~\ref{scanobj}), then mimic and incorporate these corruptions into the clean partial point clouds from the PCN dataset. We categorize the corruptions into two types: External Corruptions, originating from points belonging to other objects or the background, and Internal Corruptions, which displace or distort points from the target object (shown in Figure \ref{cpccdfig}). 
Unlike previous works \cite{pointcloudc, robo3d}, which use fixed corruption levels, we introduce randomness by allowing a range of values for the defining parameters, within spatial constraints. This design aims to reflect the unpredictable nature of noise and corruption in real-world point cloud scans, which is lacking in current completion benchmark datasets.

To address the lack of robust completion algorithms, we introduce Denoising-While-Completing (DWCNet). DWCNet utilizes a novel Noise Management Module (NMM) to classify features from a noisy partial point cloud into clean and noisy categories, alleviating the issues of outliers. Clean features are filtered through contrastive learning in feature space. NMM employs Multi-Head Self-Attention (MHSA) to determine structural relations and multi-scale convolutions to capture noise at different scales. This integrated approach ensures that the point cloud is both denoised and completed in a single, cohesive step, enhancing output quality.
Overall, our contributions are:
\begin{itemize}

\setlength{\itemsep}{-1mm}
\item We formulate and systematically approach an existing but underexplored problem in point cloud completion: the challenge of completing highly corrupted (noisy) partial point clouds.
\item We introduce a novel corrupted point cloud completion dataset (CPCCD) as the first robustness benchmark in the field of point cloud completion.
\item We offer the first systematic evaluation of the robustness of completion networks, examining how robustness relates to different types of corruptions and network architectures. 
\item We introduce DWCNet, a completion algorithm that integrates denoising and completion through a novel Noise Management Module, producing relatively clean, complete point clouds from noisy inputs. DWCNet achieves state-of-the-art results on the PCN, CPCCD, and ScanObjectNN datasets, demonstrating robustness on corrupted data.

\end{itemize}

\section{Related Work}
\label{relatedwork}

This section surveys datasets and techniques used in point cloud completion. Point clouds, essential for 3D vision and research, can be acquired through active or passive methods \cite{acq}. Active acquisition directly captures point clouds using sensors like LiDAR, Structured Light, and Time of Flight (ToF) cameras \cite{regtut}, while passive acquisition derives point clouds from other data types such as meshes, depth images, and videos \cite{acq}. Point clouds from active acquisition are more susceptible to noise and corruption due to factors like the acquisition environment, measurement devices, or human error \cite{modelnetc}, in contrast to those acquired passively.

\noindent\textbf{Datasets in Point Cloud Completion:}
Due to challenges in active acquisition, researchers often use passively acquired, clean, and synthetic pairs of partial and complete point clouds for training. Notable benchmarks include PCN \cite{pcn}, MVP \cite{vrcnet}, and Completion3D \cite{topnet}. Techniques such as back projection \cite{pcn}, adding random Gaussian noise \cite{multimodal}, and ``noised back projection'' from depth images \cite{adapointr} have been employed to make partial point clouds resemble real-world scans. However, these methods do not fully capture the noise and corruptions common in real-world data \cite{modelnetc, pointcloudc, robo3d, 3dcorrupt}. While recent work has explored corruptions in point clouds for other tasks \cite{modelnetc, pointcloudc, robo3d, 3dcorrupt}, they remain absent in point cloud completion datasets. We address this gap by introducing a robustness benchmark with corrupted and noisy partial point clouds, offering a more challenging and practical evaluation environment.

\noindent\textbf{Techniques in Point Cloud Completion:} Point cloud completion techniques can be broadly categorized into traditional and deep learning-based approaches \cite{mytvcg}. Traditional methods rely on parameter optimization to fill small missing regions \cite{mytvcg}, whereas deep learning methods use neural networks to extract features and complete larger missing areas \cite{dldl}. Recent research emphasizes improving performance through architectural innovations, including point-based, convolutional, graph-based, generative, and transformer-based models. Among these, transformer-based networks like \cite{adapointr} have shown strong results on both synthetic and real-world datasets.

Despite these advancements, most methods assume clean point clouds, overlooking the effects of noise and corruption. A few notable efforts address this issue: Agrawal et al. \cite{high} apply a denoising auto-encoder before using a network trained on clean data; Arora et al. \cite{multimodal} study noise tolerance with Gaussian noise; PDCNet \cite{pdcnet} uses density-based clustering; Ma et al. \cite{csnet} apply segmentation and Farthest Point Sampling (FPS) to exclude outliers; and Li et al. \cite{noiseaware} use outlier suppression during patch training. However, only very few approaches consider multiple noise and corruption types simultaneously. Our work evaluates various completion models under these conditions and introduces DWCNet, a robust network enhanced with a specialized module for handling corruption.

\noindent\textbf{Corruption in Point Cloud Learning tasks:}
Recently, there has been growing interest in studying dataset corruptions in 3D learning tasks. Notable efforts include ShapeNet-C \cite{pointcloudc}, ModelNet-C \cite{modelnetc}, Robo3D \cite{robo3d}, and 3D-Corruptions-AD \cite{3dcorrupt}. Ren et al. \cite{pointcloudc} introduce ModelNet-C and ShapeNet-C by applying various noise types and augmentations for classification and part segmentation. Robo3D and 3D-Corruptions-AD focus on outdoor autonomous driving scenarios, providing corrupted LiDAR scans and benchmarking models on detection and segmentation. Robo3D also introduces corrupted versions of real-world datasets: KITTI-C, Waymo-C, and nuScenes-C. In contrast, our proposed dataset, CPCCD, is designed specifically for the task of completion of point clouds of objects captured in indoor environments. Although it follows the broader trend of augmenting existing datasets, CPCCD introduces a diverse range of corruption types with randomized parameters to better reflect the unpredictable nature of real-world noise -- rather than relying on rigidly defined severity levels or fixed noise patterns.

\begin{table*}[t!]
\caption{Comparison of different completion benchmark datasets. ``Type'' shows \kl{whether} 
the datasets are synthetic or real-world and ``Paired'' shows if the datasets contain pairs of partial and complete point clouds.} 
\label{whycpccd}
\centering
\begin{scriptsize}
\small
\begin{tabular}{p{0.2\linewidth}p{0.12\linewidth}p{0.05\linewidth}p{0.55\linewidth}}
\hline
\textbf{Dataset} & \textbf{Type} & \textbf{Paired} & \textbf{Corruption} \\
\hline
PCN\cite{pcn} & Synthetic & Yes & Back-projection \\
Completion3D \cite{topnet} & Synthetic & Yes & Cropped \\
MVP \cite{vrcnet} & Synthetic & Yes & Back projection \\
ShapeNet33(54)\cite{shapenet} & Synthetic & Yes & Back projection, cropped  \\
KITTI\cite{kitti} & Real-world & No & Occlusion \\
ScanNet\cite{scannet} & Real-world & No & Occlusion \\
Projected Shapenet\cite{adapointr} & Synthetic & Yes & Back projection with Gaussian noise \\
\textbf{Ours(CPCCD)}  & Synthetic & Yes & Back projection, External object Interference  ($E_{OI}$),  Background Interference Wall, Floor ($BI_w $,  $BI_f$), Occlusion by Other Objects($O_{BOO}$), Dynamic Jitter with Trajectory ($D_{JT}$), Triaxial Rotation ($T_R $), Isometric Scaling ($I_S$), Random Combined Corruptions ($R_{CC}$)\\
\hline
\end{tabular}
\end{scriptsize}
\vspace{-0.40cm}
\end{table*}

\noindent\textbf{Point Cloud Denoising} aims to remove noise and corruption to recover clean point clouds \cite{denoisedenoise}. Methods include filter-based, optimization-based, and deep learning approaches. Most are designed for specific noise types and struggle with the complex, compound noise in real-world scans \cite{denoisedenoise}. Current state-of-the-art techniques typically denoise complete point clouds and handle only limited corruption types \cite{denoisereview}. Among the limited works that attempt to address multiple noise types are: Score-Denoise \cite{scoredenoise}, which estimates the noise distribution score of noisy point clouds; PathNet \cite{pathnet}, which introduces path-selective denoising for different noise types; NoiseTrans \cite{hou2023noisetrans}, which leverages self-attention mechanisms for point cloud denoising; and Total Denoising \cite{totaldenoising}, which adapts unsupervised image denoising techniques to 3D point clouds. While these methods show progress on specific or similar noise types, none of them address denoising for partial point clouds or diverse corruptions simultaneously, which is the focus of our work.

\noindent\textbf{Contrastive learning} is a self-supervised or supervised technique that enables a model to distinguish between similar and dissimilar samples \cite{pointcontrast, epcontrast}. It helps learn meaningful 3D representations by leveraging local geometric consistency, and has been applied in point cloud pretraining \cite{du2024pcl}, as well as improving performance in point cloud detection \cite{votenet} and segmentation \cite{pointcontrast}. 

\section{Corrupted Point Cloud Completion Dataset (CPCCD)}
\label{sec:cpccd}

As identifying clean vs. noisy points at a fine level is infeasible \cite{denoisereview}, both completion and denoising methods typically rely on synthetic datasets for training and evaluation \cite{denoisereview}. However, current completion benchmark datasets \cite{pcn, vrcnet, topnet} exhibit little to no corruption. Real-world datasets like KITTI \cite{kitti} and ScanNet \cite{scan2cad} are preprocessed before evaluation \cite{p2c, robo3d}. The domain gap between real and synthetic data complicates fair evaluation \cite{p2c}, and qualitative results are often subjective. Due to the lack of ground truth and the unpredictable nature of real-world corruptions, real data cannot serve as a reliable robustness benchmark. Our goal is to create a robustness benchmark for object-level point cloud completion by simulating corruptions likely to occur in indoor environments. 

To create CPCCD (Figure~\ref{cpccdfig}), we first visually inspect a real-world indoor point cloud dataset, ScanObject-NN \cite{scanobjectnn} (shown in Figure \ref{scanobj}), to identify common corruptions in indoor environments. We selected ScanObject-NN because it covers all objects in the ScanNet dataset, rather than just those chosen by researchers in completion studies \cite{p2c, realdiff}. After identifying these corruptions, we apply them to the partials in the PCN dataset~\cite{pcn}.

We categorized the observed corruptions into two groups: External and internal corruptions. \textbf{External corruptions} add points from unrelated objects or backgrounds (e.g., floors, walls), while \textbf{Internal corruptions} disrupt or displace points within the target object, often due to sensor errors, occlusions, or scaling issues. We applied these corruptions to the partials in the PCN dataset \cite{pcn}, preserving the train, test, and validation splits. The PCN dataset contains partial and complete point clouds for 30,974 shapes from 8 categories (Plane, Cabinet, Car, Chair, Lamp, Couch, Table, Boat), derived from ShapeNet \cite{shapenet}. Complete point clouds are generated by uniformly sampling ShapeNet CAD models, and partials are created by back-projecting depth images from 8 viewpoints. We then applied the observed eight different corruptions (discussed below) to each of the partials. CPCCD consists of the equivalent of 8 PCN datasets, with 8 variations for each of the point clouds in the training, testing and validation dataset. A breakdown about the number of point clouds in CPCCD is given in \ref{appendix1}. We compare CPCCD to other completion datasets in Table \ref{whycpccd}.
 
\begin{table*}[t!]
\caption{CPCCD dataset and its parameter definition. Possible Severity level ($P$) represent all the possible combinations of parameter values that can determine the severity level of the corruption.}
\label{cpccdtable}
\Large
\begin{scriptsize}
\begin{tabular}{|p{0.05\linewidth}|p{0.12\linewidth}|p{0.22\linewidth}|p{0.3\linewidth}|p{0.15\linewidth}|}
\hline
\textbf{Noise} & \textbf{Noise subclass} & \textbf{Purpose} & \textbf{Parameter Domain} & \textbf{Possible Severity Levels} \\
\hline
External &  External Object Interference $(E_{OI})$ & Represents corruption by points belonging to the other objects & \raggedright $No\_of\_Objects (N_o)=[1,2,3]$, \newline$No\_of\_points (N_p)= [1/16,1/12,1/8,1/4]$, 
\newline$No\_of\_Shapes (N_s)= [12]$,  
\newline$Distance\_from\_obj (Nd) = [0.05,0.2]$ & $ P_{E_{OI}} = N_o  \times  N_p  \times  N_s  \times  N_d $ \\
\hline
External &   Background Interference: Wall $ (BI_w)$ & Represents points from walls or vertical flat surfaces captured in the point cloud scan of the object &   Distance from the object, $ : N_d = [0.01 - 0.05] $ &  $ P_{BI_{W}} = N_d$ \\
\hline
External &  Background Interference: Floor $ (BI_f)$ & Represents the floor points & $ - $ &  $P_{BI_{F}} = 1 $ \\
\hline
Internal &  Occlusion by Other Objects $(O_{BOO})$ & Represents missing points due to occlusions by other objects in the scene & \raggedright $No\_of\_Objects(N_o) = [1,2,3,4] $,  \newline$No\_of\_points(N_p) = [1/8,1/7,1/6, 1/5,1/4,1/3]$, \newline$No\_of\_Shapes(N_s) = [12]$, \newline$Distance\_from\_obj(N_d) = [0.05,0.2] $ & $ P_{O_{BOO}} = N_o \times N_s \times N_d \times N_s $ \\
\hline
Internal &  Dynamic Jitter with Trajectory $ (D_{JT}) $ & represents displacement of points in the object scan in the trajectory &  Jitter amount$(J_a) = [0.01, 0.05] $, \newline Trail Distance$(T_d) = [0.02, 0.04] $ & $ P_{D_{JT}} = J_a \times T_d $ \\
\hline
Internal &  Triaxial Rotation $ (T_R) $ & represents change in viewpoints due to environmental factors &  $\theta_x , \theta_y, \theta_z $ &  $  P_{T_{R}} =\theta x \times \theta y \times \theta z $ \\
\hline
Internal &  Isometric Scaling $ (I_S) $ &  represents difference of scale between real world objects &  $ N_s = [0.25, 2] $  &  $ P_{I_S} = N_s $\\
\hline

Random &  Random Combined Corruption  $(R_{CC}) $ & represents the various combinations that corruptions appear appear in the real world & a random combination
of any 2 to 7 of the corruptions above $ \{ S \subseteq C \mid 2 \leq |S| \leq 7 \} $ & $ P_{R_{CC}} = P_{E_{OI}} \times P_{BI_{W}} \times P_{BI_{F}} \times P_{O_{BOO}} \times P_{D_{JT}} \times P_{T_{R}} \times P_{I_S} $ \\

\hline
\end{tabular}
\end{scriptsize}
\vspace{-4mm}
\end{table*}

\noindent\textbf{External Object Interference $(E_{OI})$:} To generate external object corruption, we randomly select one to three shapes from a set of 12 basic shapes, $N_s=12$: Circle, Square, Rectangle, Triangle, Ellipse, Hexagon, Diamond, Parallelogram, Cylinder, Sphere, Cube, and Pyramid. These shapes were chosen as they can represent parts of more complex objects. Simulating the scenario where external objects are captured in the point cloud scan of the target object (like those in Figure \ref{scanobj}), we randomly apply $N_o = [1,2,3]$ shapes. We control their proximity to the target object using $N_d$, which ranges from $N_d = [0.05, 0.2]$ to ensure that the shapes do not occlude the object but remain within its bounding box. The added shapes contribute $N_p$ points, where $N_p = [1/16, 1/12, 1/8, 1/4]$, ensuring that the corruption introduces a meaningful challenge rather than simple outliers that could be removed by basic outlier removal techniques.

\noindent\textbf{Background Interference: Wall $(BI_w)$}: This simulates the presence of walls, doors, and other vertical, flat objects typically found in indoor scenes. The wall is positioned behind the target object, so points from the wall appear within the object's bounding box without occluding it. The proximity is controlled by $N_d$, where $N_d = [0.01, 0.05]$. Points on the wall that are occluded by the object are removed by projecting the object’s points onto the wall and eliminating overlapping points.

\noindent\textbf{Background Interference: Floor $(BI_F)$}: This corruption simulates the floor and appears only when the bottom quarter of the object is captured in the point cloud. Points from the floor that may be occluded by the object are removed by projection.

\noindent\textbf{Occlusion by Other Objects $(O_{BOO})$}: This corruption simulates occlusion by other objects in the scene. It removes points from the object of interest based on simple shapes $N_s = 12$ (as described in external object interference $(E_{OI})$). $N_p$ represents the number of points dropped due to occlusion, with values $N_p = [1/8, 1/7, 1/6, 1/5, 1/4, 1/3] * N_t$, where $N_t$ is the total number of points in the object. This range ensures a meaningful portion of the object remains while causing a significant change.

\noindent\textbf{Dynamic Jitter with Trajectory $(D_{JT})$}: This corruption simulates jitter caused by motion. Unlike other corruption datasets \cite{pointcloudc}, we apply jitter only to a select few points in the object, with displacement $J_\alpha$ ranging from $[0.01, 0.05]$. These points are then displaced along a trail distance $T_d = [0.02, 0.04]$ meters in a random direction. The small trail distance reflects an indoor environment scenario. 

\noindent\textbf{Isometric Scaling $(I_s)$}: This corruption simulates scale mismatches between the input and ground truth, as well as real-world size differences between objects (e.g., a chair vs. a car). To account for this, we vary the scale of the partials by a factor $I_s$ within the range $[0.25, 2]$.

\begin{figure*}[th!]
\centerline{\includegraphics[width = \linewidth]{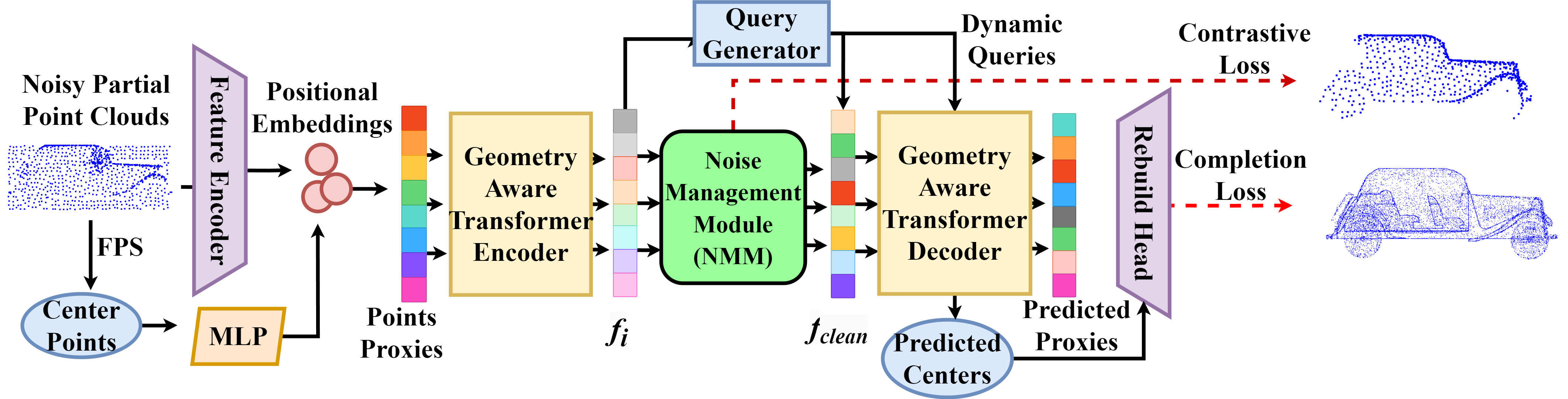}}
\caption{Overall architecture of DWCNet. The input point cloud is first down-sampled and local features are extracted using DGCNN~\cite{dgcnn}, following the backbone of AdaPoinTr~\cite{adapointr}. The extracted features, combined with positional embeddings, are passed through a transformer encoder. The resulting encoded features are then processed by the NMM, which classifies them as either “clean” or “noisy.” Only the clean features are passed to the transformer decoder, which predicts a coarse point cloud and point proxies. These are further refined and upsampled to a complete point cloud by using a folding decoder.}
\label{dwcnetfig}
\vspace{-0.1 in}
\end{figure*} 

\noindent\textbf{Triaxial Rotation $(T_R)$ }: Rotation accounts for any misalignment during acquisition. We rotate the angles of the $x, y,$ and $z$ axes by $\theta_x , \theta_y, \theta_z $ which can range from zero to ten degrees each.

\noindent\textbf{Randomly Combined Corruption $(R_{CC})$ }: This corruption involves a random combination of two or more of the seven corruptions described above. It simulates how real-world corruptions often occur in an unpredictable and overlapping manner. Let \( C \) be the set of all possible corruptions: $C = [E_{OI}, BI_W, BI_F, O_{BOO}, D_{JT}, I_S, T_R]$.
The random corruption \( R_{CC} \) is any subset \( S \) of \( C \) such that:
$R_{CC} \in \{ S \subseteq C \mid 2 \leq |S| \leq 7 \}$.
All corruptions and their levels are described in Table \ref{cpccdtable}. Examples of objects from the CPCCD dataset are shown in Figure \ref{cpccdfig}.  

\section{Denoising-While-Completing Network (DWCNet)}
\label{dwcnet}
Our robustness analysis on the proposed CPCCD dataset (Section~\ref{robo_section}) shows that state-of-the-art completion models are highly vulnerable to noisy inputs. Although retraining on CPCCD improves their robustness, these models still struggle to effectively remove noise. This limitation motivates our proposed solution: the Denoising-While-Completing Network (DWCNet). DWCNet tackles the limitation by integrating a dedicated \textbf{Noise Management Module (NMM)} to a base completion algorithm to simultaneously denoise and complete corrupted partial point clouds. The key idea behind DWCNet is to identify and filter out noisy features in the pipeline, allowing the model to concentrate on clean, informative data during completion. This selective processing reduces noise propagation and leads to more accurate and robust reconstructions. 

In our design, we choose AdaPoinTr~\cite{adapointr} as the base completion algorithm
due to its integrated Gaussian denoising mechanism and strong performance on CPCCD (see Section~\ref{robo_section}). We then introduce the NMM module in its core to separate clean and noisy features (Figure~\ref{dwcnetfig}).

Specifically, the DWCNet pipeline begins with an input point cloud \( P = \{P_x, P_y, P_z\} \) of size \( N \times 3 \), where \( N < 16384 \). This input is downsampled to 2048 points and grouped into local patches (e.g. 32 groups of 64 points), which are embedded into higher-dimensional representations -- point proxies -- using multilayer perceptrons (MLPs) and positional encodings, resulting in a feature tensor of shape \( 32 \times 256 \). These proxies are processed by a geometry-aware Transformer encoder that preserves local structure. The resulting global feature \( f_i \) is passed to the Noise Management Module (NMM), which separates it into clean (\( f_{clean} \)) and noisy (\( f_{noisy} \)) components, padding as needed to maintain dimensional consistency. The clean features \( f_{clean} \) replace \( f_i \) for subsequent processing. An adaptive query generation module then produces 1024 learned queries of shape \( 1024 \times 256 \), which are refined by a Transformer decoder. The decoder output is mapped to 3D coordinates via a point generation head, yielding a coarse completed point cloud of shape \( 1024 \times 3 \). This is then upsampled to a final point cloud of shape \(16384 \times 3\).

\subsection{Noise Management Module (NMM)}

Inspired by speech enhancement research \cite{speech,speech2,speech3}, which jointly models and separates signal and noise, we propose the Noise Management Module (NMM). It is a dual-path neural network that classifies features encoded by the completion network as noisy or clean using self-attention and contrastive learning \cite{contrastive}. The clean features are then passed back into the completion pipeline to predict the final complete point clouds.

\begin{figure}[t!]
\centerline{\includegraphics[width = \linewidth]{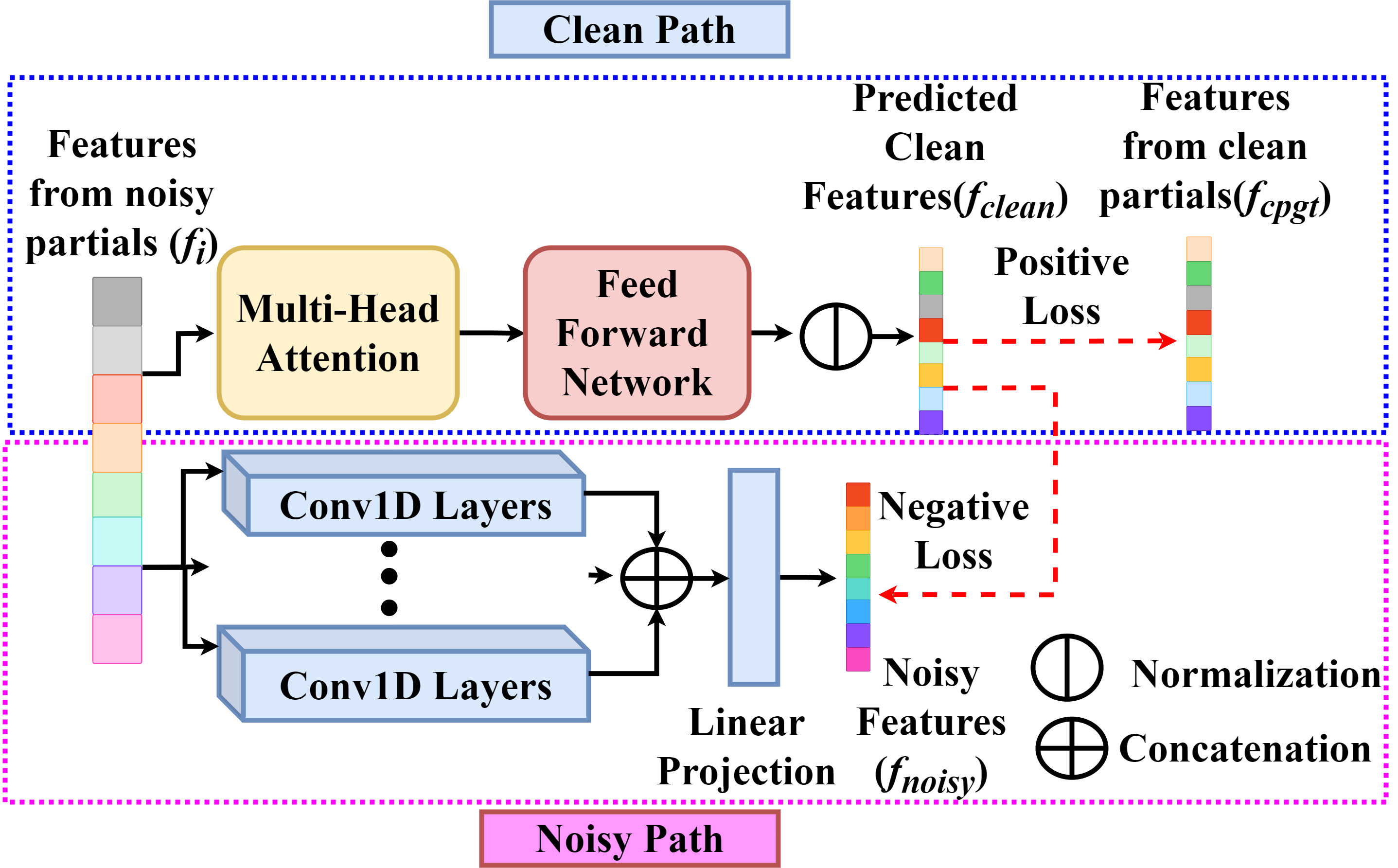}}
\vspace{-0.1 in}
\caption{Noise Managment Module (NMM)}
\label{nmmfig}
\vspace{-0.1 in}
\end{figure} 

The architectural details of NMM are as follows. NMM processes input features through two parallel pathways. The first path is a \textit{clean path}, where a transformer-based module captures long-range dependencies via multi-head self-attention \cite{mhsa}, followed by a position-wise feed-forward network (FFN). Pre-Layer normalization is applied to prevent feature magnitude divergence and residual connections are used to mitigate vanishing gradients. The self-attention employs eight attention heads to capture global dependencies while the FFN refines features per-position. Given an input feature $f_i$, \[
f_i \in \mathbb{R}^{B \times L \times D}
\] where $B$ is the batch size, $L$ is the sequence length $D$ is the feature dimension, and $+$ represents addition.

\begin{align}
    f_{\text{MHSA}} &= \text{LayerNorm}(f_i + \text{MHSA}(f_i))\\
f_{clean} &= \text{LayerNorm}(f_{\text{MHSA}} + \text{FFN}(f_{\text{MHSA}}))
\end{align}
where MHSA represents multi-head self-attention and LayerNorm represents normalization. This can be summarized as: 
\begin{equation}
    f_{clean} = f_i + \underbrace{\text{MHSA}(f_i)}_{\text{Global context}} + \underbrace{\text{FFN}(f_i + \text{MHSA}(f_i))}_{\text{Feature refinement}}
\end{equation}

The second path is a \textit{noisy path} that uses multi-scale convolutions, consisting of multiple parallel 1D convolutions, to capture noise at different scales. The process is summarized in the equation below.

\begin{equation}
f_{\text{noisy}} = W_m \cdot \text{Concat}\big[\{\text{Conv}_k(f_i^T)\}_{k \in \{1,3,5\}}\big]
\end{equation}
where $f_{noisy}$ is noisy features, $W_m\in \mathbb{R}^{D \times 3D}$ is learnable merge weights, $\text{Conv}_k$ is 1D convolution with kernel size $k$, $f_i^T \in \mathbb{R}^{B \times D \times L}$ is transposed input feature. $k$ represents convolution kernel sizes of $\{1,3,5\}$. $B$ is the batch size, $L$ is the sequence length and $D$ is the feature dimension (1024 in our case).

The features $f_{clean}$ and $f_{noisy}$ from the clean and noisy paths are then linearly projected and merged. The dual-path architecture of the NMM enables explicit separation of processing pathways, allowing specialized feature learning for clean and noisy components. The full NMM architecture is shown in Figure \ref{nmmfig}.

The NMM learns discriminative features through a contrastive learning approach, maximizing similarity between clean features and ground truth while minimizing similarity between clean and noisy features. In the NMM, contrastive learning is implemented as follows: First, all input features from the corrupted partial, already processed by the transformer encoder, $f_{i}$, clean partial ground truth $f_{cpgt}$, and complete ground truth point clouds $f_{gt}$ are normalized along the feature dimension. After normalization, two losses are used to optimize feature separation. The first is the \textit{Positive Loss}, which measures the cosine similarity between clean features $f_{clean}$ and the clean partial ground truth$f_{cpgt}$. Minimizing this loss maximizes similarity between the clean features $f_{clean}$ and the clean partial ground truth $f_{cpgt}$. The second is the \textit{Negative Loss}, which computes all pairwise similarities between the clean $f_{clean}$ and noisy features$f_{noisy}$. Diagonal masking is applied to avoid self-comparisons, and temperature scaling \cite{tempscale} is used for stable optimization. Temperature scaling helps balance the positive and negative terms: a lower temperature $t=0$ exaggerates the difference between clean and noisy features, while a higher $t$ (e.g., 1.0) leads to a more uniform separation. For contrastive learning, we use a temperature scaling \(t = 1\) following findings from our ablation study (see Section \ref{ablationsection}).

Mathematically, the losses are defined as follows. Given normalized features from the clean partial point cloud \(f_{\text{cpgt}} \in \mathbb{R}^{B \times L \times D}\), the clean complete ground truth \(f_{\text{gt}},\in \mathbb{R}^{B \times L \times D}\), the noisy partial input \(f_{\text{input}} \in \mathbb{R}^{B \times L \times D}\), the clean features from the clean path  \(f_{\text{clean}},\in \mathbb{R}^{B \times L \times D}\) and the temperature parameter $ t = 1 $, the Positive Loss is defined as follows. 

\begin{equation}
    \mathcal{L}_{\text{pos}} = -\frac{1}{B \cdot L} \sum_{i=1}^{B} \sum_{j=1}^{L} f_{\text{clean}}^{(i,j)} \cdot f_{\text{cpgt}}^{(i,j)}
\end{equation}
where $\cdot$ denotes the dot product. $B$ is the batch size, $L$ is the sequence length and $D$ is the feature dimension(1024). The \textit{Negative Loss} is defined as follows.
\begin{equation}
    \mathcal{L}_{\text{neg}} = t \cdot \log \sum_{\substack{i=1, j=1, i \neq j}}^{B \cdot L} \exp \left( \frac{{f}_{\text{clean}}^{(i)} \cdot f_{\text{noisy}}^{(j)}}{t} \right)
\end{equation}
\noindent The total combined loss $L_{NMM}$ is:
\begin{equation}
    \mathcal{L}_{\text{NMM}} = \mathcal{L}_{\text{pos}} + \mathcal{L}_{\text{neg}}
\end{equation}
\noindent This contrastive loss from NMM is then added to completion loss for an end-to-end training.

\begin{equation}
    \mathcal{L}_{\text{Total}} = \mathcal{L}_{\text{completion}} + \mathcal{L}_{\text{NMM}}
\end{equation}


\section{Robustness Evaluation}
\label{robo_section}
The following experiments simulate the scenario where existing techniques, trained on the clean PCN dataset, are tested on the corrupted CPCCD dataset. These models are then fine-tuned and retested on CPCCD. We now discuss the experiment setup, completion algorithms, and evaluation metrics.

\subsection{Experimental Setup}
\label{setup}

Our primary selection criterion for completion models is the network architecture. We aimed to test at least one algorithm from all architecture categories: Convolution-based, point-based, graph-based, transformer-based, and generative model-based (Generative Adversarial Network (GAN)-based, Variational AutoEncoder (VAE)-based, and generative-diffusion model-based). It is important to note that most completion networks are hybrids of various architectures. Although we tried to include models from all categories, the availability of source codes and pre-trained models influenced our final selection. We prioritized networks that use the PCN dataset for evaluation, as CPCCD is based on this dataset.

We selected the following completion networks: point-based encoder-decoder (PCN \cite{pcn}), graph-based (FoldingNet \cite{foldingnet}), convolution-based (GRNet \cite{grnet}), and transformer-based (PoinTr \cite{pointr}, SnowflakeNet \cite{snowflakenet}, and AdaPoinTr \cite{adapointr}). We evaluated the selected algorithms using their provided pretrained weights on the different corruptions of the CPCCD dataset, with results shown in Tables \ref{resulttable1}, \ref{resulttable1fscore}, \ref{resulttable1fidelity} and \ref{resulttable1cdl2}. Then, we fine-tuned these models on the Rcc category of the CPCCD dataset and evaluated them again, as shown in Tables \ref{resulttable2}, \ref{resulttable2fscore}, \ref{resulttable2fidelity} and \ref{resulttable2cdl2}. All models were trained and fine-tuned using an NVidia A100 GPU (40GB VRAM), i9-12900K Intel(R) CPU, following the procedures outlined in the original papers. 

\subsection{Evaluation Metrics}
We employ Chamfer Distance, F-score and Fidelity, which are commonly used by state-of-the-art completion algorithms, as our performance metrics. 

\noindent\textbf{Chamfer Distance (CD) \cite{cd}:} represents the average distance of closest point between two point clouds. Given two sets of points $D_1$ and $D_2$, the Chamfer Distance between the prediction set $D_1$ and the ground truth $D_2$ is defined as:  
\small
\vspace{-0.15cm}
\begin{equation} \label{eq:veclen} 
\begin{aligned}[b]
CD(D_{1}, D_{2}) = \dfrac{1}{|D_{1}|} \sum_{x \in D_{1}}^{} min_{y \epsilon D_{2}}\left \| x - y  \right \|_{2} + \\
\frac{1}{|D_{2}|} \sum_{y \in D_{2}}^{} min_{x \epsilon D_{1}}\left \| y - x  \right \|_{2}
\end{aligned}
\end{equation}
\vspace{-0.1 in}
\normalsize

\noindent where $\left \| \cdot \right \|_{2}$ is the Euclidean distance. Chamfer Distance is used both as an evaluation measure and a loss function for optimizing learning-based algorithms. It is the most popular metric used in completion. However, it can be insensitive to local density distribution and outliers \cite{dcd}. The lower the Chamfer Distance, the better the performance. For our experiment, we use CD-L1 which uses L1-norm to calculate difference between two points. L1-norm calculates the sum of the absolute values of the vector elements \cite{CDL1CDL2}. 

\noindent\textbf{F-score \cite{fscore}}:
measures completion accuracy through precision and recall. Given the predicted point cloud ($D_1$) and ground truth ($D_2$), the F-score is defined as: 
\small
\begin{equation} \label{eq:veclen5} 
F\textrm{-}score(\delta ) = \frac{2G(\delta)H(\delta)}{G(\delta)+H(\delta)}
\end{equation}
\normalsize
\noindent where $G(\delta)$ and $H(\delta)$ represent point-wise precision and recall for threshold $\delta$. The higher the F-score the better.

 
\noindent\textbf{Fidelity \cite{pcn}}: measures the preservation of input data. It computes the average distance from each point in the input to its nearest neighbor in the output. Lower fidelity indicates superior performance, as it means the output more closely matches the input. 

\section{Results and Robustness Analysis}
\label{sec:analysis}

\begin{figure*}[ht!]
\centerline{\includegraphics[width= 0.95\linewidth]{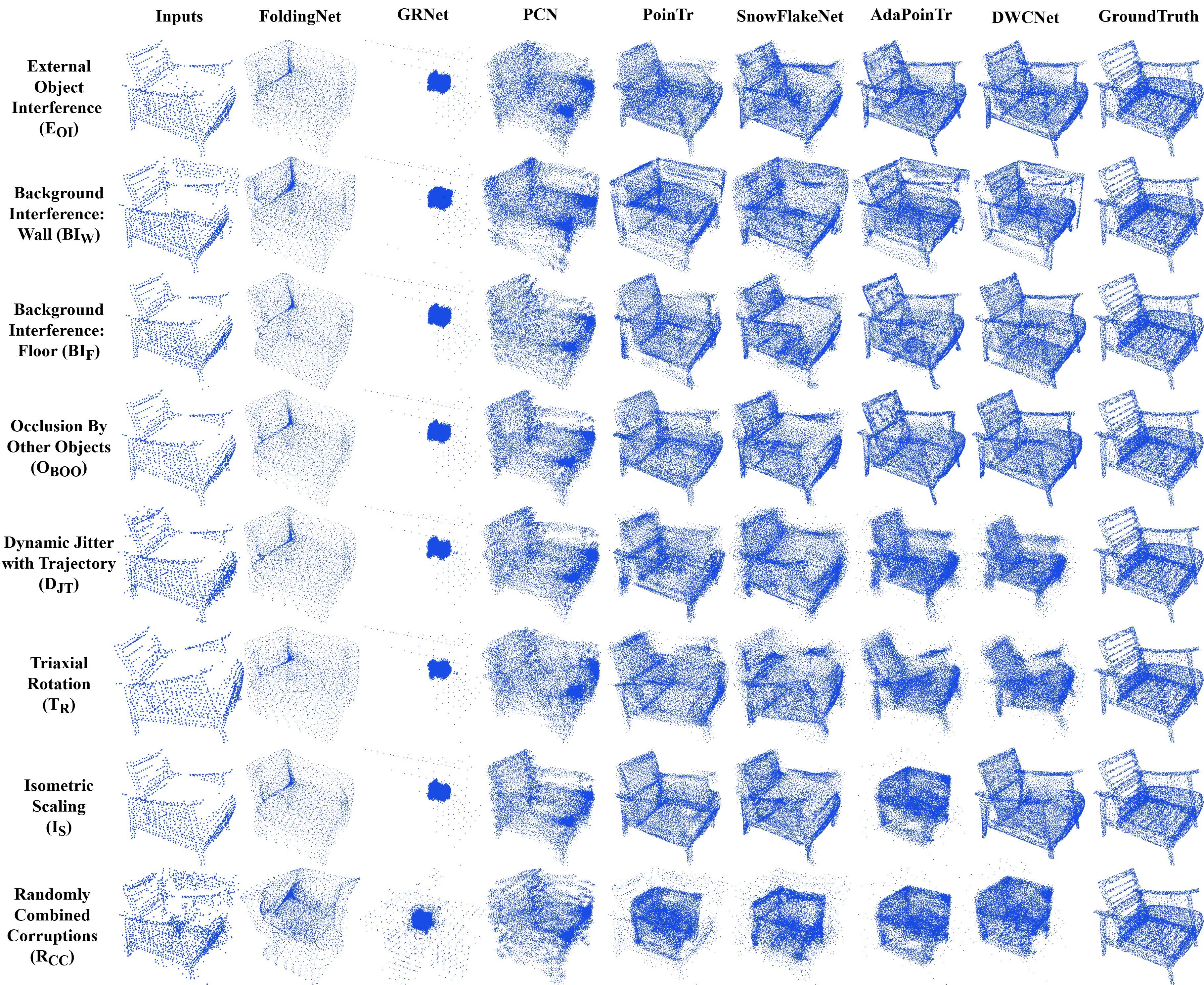}}
\caption{Example of results on CPCCD dataset before fine-tuning: chair}
\label{chairfigs}
\vspace{-0.1 in}
\end{figure*}

\begin{figure*}[ht!]
\centerline{\includegraphics[width= 0.95\linewidth]{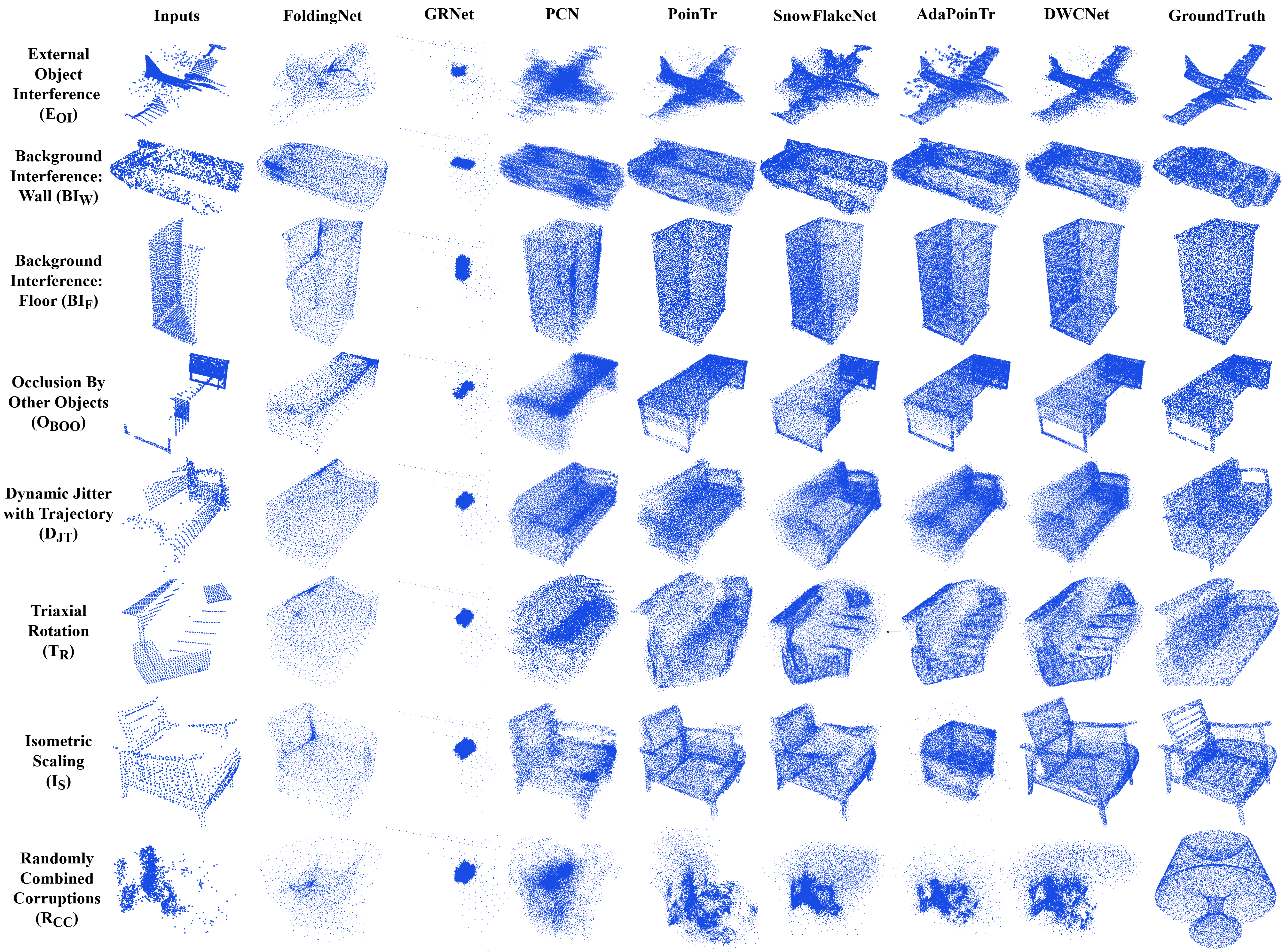}}
\caption{Example of results on CPCCD dataset before fine-tuning: multiple categories}
\label{bffigs}
\vspace{-0.1 in}
\end{figure*}

The first experiment was conducted to evaluate the ability of completion networks to generalize to corrupted data. In this section, we make observations and analyze the results.

\subsection{Robustness Before Fine-tuning}
The $CD_{L1}$, $Fscore$ and $fidelity$ results before fine-tuning are shown in Tables \ref{resulttable1}, \ref{resulttable1fscore} and \ref{resulttable1fidelity} respectively.

\begin{table}[h!]
    \caption{Completion Results on CPCCD datasets. The result shows $CD_{L1}$ results of the completion. The lower the $CD_{L1}$ the better the result. The superscripts 'PB', 'GB', 'CB' and 'TB' stand for Point-Based, Graph-Based, Convolution-Based and Transformer-Based respectively.} 
    \label{resulttable1} 
    \begin{scriptsize}
    \centering
    \setlength{\tabcolsep}{1.4pt}
    {\small
\resizebox{\linewidth}{!}{%
    {\begin{tabular}{c c c c c c c c c c c }
        \toprule 
        Network (CDL1$\downarrow$) & clean & $E_{OI}$ & $ {BI_W} $ & $ BI_F $ & $ O_{BOO} $ & $D_{JT}$ & $T_R$ & $I_S$ & $R_{CC}$ \\
        \midrule
        FoldingNet\textsuperscript{GB} & 26.493 & 33.135 & 33.135 & 42.259 & 26.469 & 27.007 & 29.417 & 44.834 & 51.694 \\
        PCN\textsuperscript{PB} & 33.148 & 40.046 & 50.847 & 38.985 & 33.152 & 34.347 & 34.8 & 45.107 & 58.376 \\
        GRNet\textsuperscript{TB} & 24.686 & 25.794 & 48.249 & 34.961 & 25.248 & 24.731 & 28.295 & 44.795 & 63.516 \\
        PoinTr\textsuperscript{TB} & 8.388 & \textbf{13.516} & 25.481 & \textbf{14.836} & 8.483 & \textbf{10.837} & \textbf{14.699} & 24.302 & 35.63 \\
        SnowFlakeNet\textsuperscript{TB} & 8.18 & 15.181 & 26.731 & 18.728 & 8.256 & 10.85 & 14.854 & \textbf{21.843} & \textbf{35.424} \\
        AdaPoinTr\textsuperscript{TB} & 6.516 & 14.764 & \textbf{24.994} & 16.167 & 6.576 & 11.68 & 15.916 & 23.016 & 36.682 \\
        DWCNET\textsuperscript{TB} & \textbf{6.493} & 14.819 & 25.648 & 16.036 & \textbf{6.531} & 11.919 & 15.915 & 273.589 & 57.919 \\
        \hline 
    \end{tabular}    
    }}}
    \end{scriptsize}
    \vspace{-0.1 in}
\end{table}

\begin{table}[h!]
    \caption{Completion Results on CPCCD datasets. The result shows $F_{Score}$ results of the completion. The higher the $F_{Score}$ the better the result. The superscripts 'PB', 'GB', 'CB' and 'TB' stand for Point-Based, Graph-Based, Convolution-Based and Transformer-Based respectively.} 
    \label{resulttable1fscore} 
    \begin{scriptsize}
    \centering
    \setlength{\tabcolsep}{1.4pt}
    {\small
\resizebox{\linewidth}{!}{%
    {\begin{tabular}{c c c c c c c c c c c }
        \toprule 
        Network ($Fscore\uparrow$) & clean & $E_{OI}$ & $ {BI_W} $ & $ BI_F $ & $ O_{BOO} $ & $D_{JT}$ & $T_R$ & $I_S$ & $R_{CC}$ \\
        \midrule
        FoldingNet\textsuperscript{GB} & 0.213 & 0.109 & 0.109 & 0.101 & 0.214 & 0.176 & 0.131 & 0.083 & 0.052 \\
        PCN\textsuperscript{PB} & 0.244 & 0.136 & 0.115 & 0.178 & 0.242 & 0.200 & 0.184 & 0.133 & 0.072 \\
        GRNet\textsuperscript{TB} & 0.273 & 0.267 & 0.150 & 0.183 & 0.260 & 0.261 & 0.174 & 0.123 & 0.083 \\
        PoinTr\textsuperscript{TB} & 0.754 & 0.592 & 0.477 & 0.611 & 0.747 & \textbf{0.634} & \textbf{0.471} & \textbf{0.398} & \textbf{0.280} \\
        SnowFlakeNet\textsuperscript{TB} & 0.741 & 0.573 & 0.485 & 0.553 & 0.738 & 0.623 & 0.438 & 0.375 & 0.251 \\
        AdaPoinTr\textsuperscript{TB} & 0.845 & \textbf{0.628} & \textbf{0.545} & 0.654 & 0.842 & 0.620 & 0.438 & \textbf{0.398} & 0.253 \\  
        DWCNET\textsuperscript{TB} & \textbf{0.847} & 0.625 & 0.532 & \textbf{0.666} & \textbf{0.845} & 0.609 & 0.439 & 0.385 & 0.254 \\ 
        \hline 
    \end{tabular}    
    }}}
    \end{scriptsize}
\vspace{-0.15in}
\end{table}

\begin{table}[h!]
    \caption{Completion Results on CPCCD datasets. The result shows $fidelity$ results of the completion. The lower the $fidelity$ the better the result. The superscripts 'PB', 'GB', 'CB' and 'TB' stand for Point-Based, Graph-Based, Convolution-Based and Transformer-Based respectively.} 
    \label{resulttable1fidelity} 
    \begin{scriptsize}
    \centering
    \setlength{\tabcolsep}{1.4pt}
    {\small
\resizebox{\linewidth}{!}{%
    {\begin{tabular}{c c c c c c c c c c c }
        \toprule 
        Network ($fidelity\downarrow$) & clean & $E_{OI}$ & $ {BI_W} $ & $ BI_F $ & $ O_{BOO} $ & $D_{JT}$ & $T_R$ & $I_S$ & $R_{CC}$ \\
        \midrule
        FoldingNet\textsuperscript{GB} & 6.525 & 6.548 & 6.548 & 11.509 & 6.507 & 5.794 & 6.469 & 7.277 & 9.974 \\
        PCN\textsuperscript{PB} & 10.373 & 10.634 & 17.271 & 12.568 & 10.365 & 9.916 & 9.977 & 11.226 & 17.032 \\
        GRNet\textsuperscript{TB} & 4.945 & 5.252 & 15.174 & 8.357 & 5.074 & 4.785 & 5.216 & 10.233 & 22.962 \\
        PoinTr\textsuperscript{TB} & 0.179 & \textbf{0.689} & \textbf{4.75} & \textbf{1.392} & 0.183 & \textbf{0.326} & \textbf{0.474} & 1.868 & 6.419 \\
        SnowFlakeNet\textsuperscript{TB} & 0.176 & 1.241 & 5.77 & 2.619 & 0.181 & 0.353 & 0.535 & 1.715 & 6.687 \\
        AdaPoinTr\textsuperscript{TB} & \textbf{0.093} & 1.096 & 5.215 & 2.222 & \textbf{0.096} & 0.399 & 0.51 & \textbf{1.344} & \textbf{5.859} \\ 
        DWCNET\textsuperscript{TB} & 0.098 & 1.082 & 5.226 & 2.28 & 0.099 & 0.409 & 0.509 & 31652.809 & 485.849 \\
        \hline 
    \end{tabular}    
    }}}
    \end{scriptsize}
\vspace{-0.15 in}
\end{table}

All evaluated networks were highly vulnerable to external corruptions, particularly external object interference ($E_{OI}$) and background interference ($BI_w$ and $BI_F$). While most algorithms showed relative robustness against internal corruptions like dynamic jitter with trajectory ($D_{JT}$) and occlusion by other objects ($O_{BOO}$), they performed poorly against isometric scaling ($I_S$) and triaxial rotation ($T_R$).

Among all corruptions, Randomly Combined Corruptions ($R_{CC}$) resulted in the highest Chamfer Distance losses, indicating the worst performance across all categories. Following $R_{CC}$, the most detrimental corruption was Background Interference: wall ($BI_w$), which covers a large portion of the bounding box, often causing significant misrepresentation (e.g., Figure \ref{chairfigs}, second column, rows two to six shows a sofa instead of a chair). Similar issues were observed with background interference floor ($BI_F$), where the presence of the floor negatively affected the completion of the chair’s legs (Figure \ref{chairfigs}, third column, rows four to seven). Isometric scale corruption, reflecting natural scale differences in real-world objects and sensor types, did not affect local details but significantly impacted performance, as algorithms trained on Chamfer Distance struggled with out-of-distribution scales. Dynamic jitter with trajectory ($D_{JT}$) produced noisy outputs, while triaxial rotation ($T_R$) showed relatively better qualitative results despite higher Chamfer Distance losses.

From the qualitative evaluation shown in Figures \ref{chairfigs} and \ref{bffigs}, we see that GRNet has the worst results (see Figure \ref{01grnet}). We hypothesize that this is due to the gridding loss function used to train GRNet. Further tests and analysis is needed to pinpoint how exactly the level and type of corruption relates with different training loss functions. We found that point-based networks, FoldingNet and PCN were able to adjust to the isometric scaling ($ I_S$) and triaxial rotation ($ T_R)$, but failed at capturing the details, whereas
transformer based networks were good at preserving detail but failed to adjust scale as required. DWCNET outperformed other models on both the clean and $O_{OBOO}$ corruption categories, setting a new benchmark as the state-of-the-art on the PCN dataset.

\subsection{Robustness After Fine-tuning}

\begin{figure*}[ht!]
\centerline{\includegraphics[width= 0.95\linewidth]{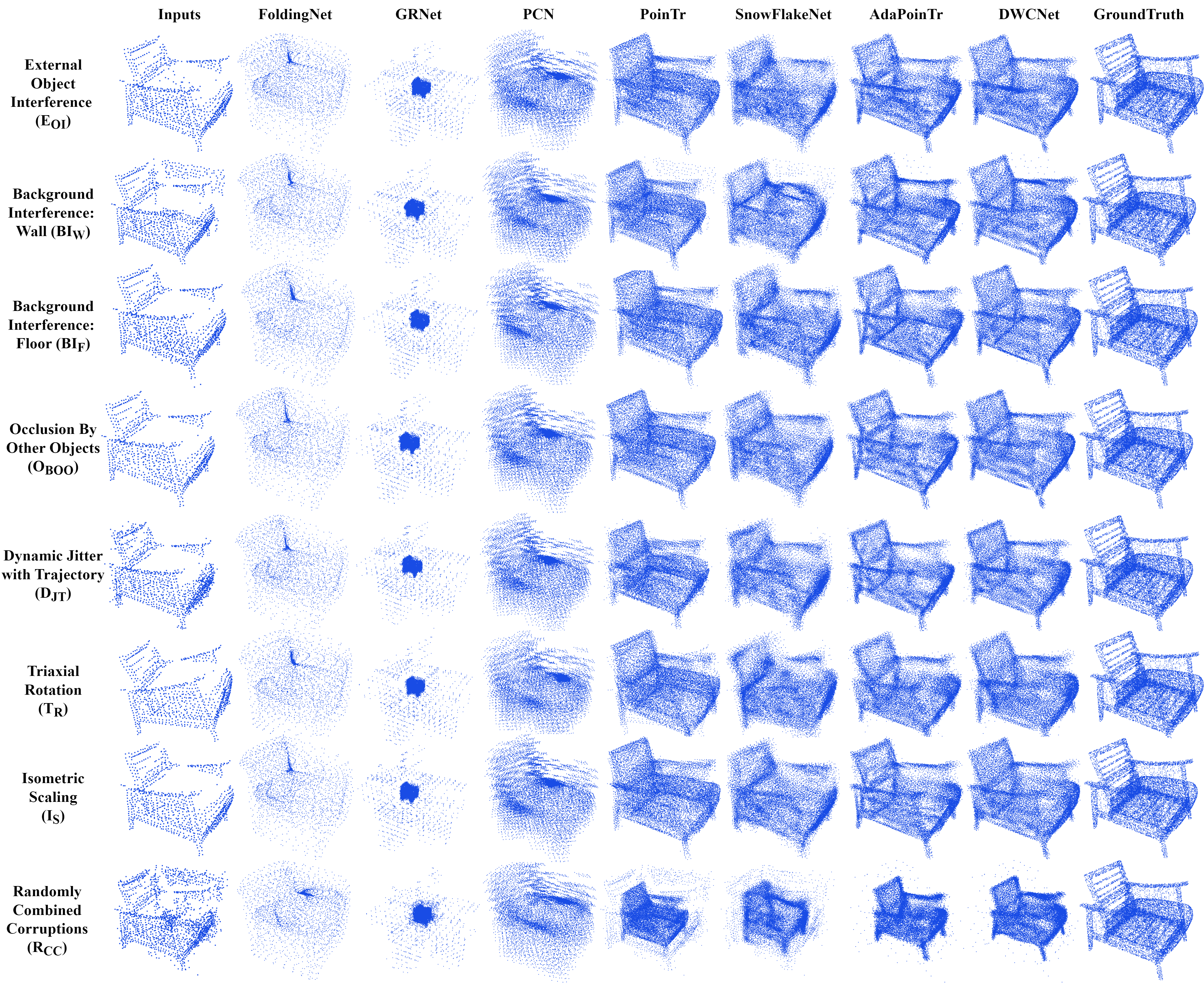}}
\caption{Example of results on CPCCD dataset after fine-tuning: chair }  
\label{chairfigsafter}
\vspace{-0.1 in}
\end{figure*}

\begin{figure*}[ht!]
\centerline{\includegraphics[width=0.95\linewidth]{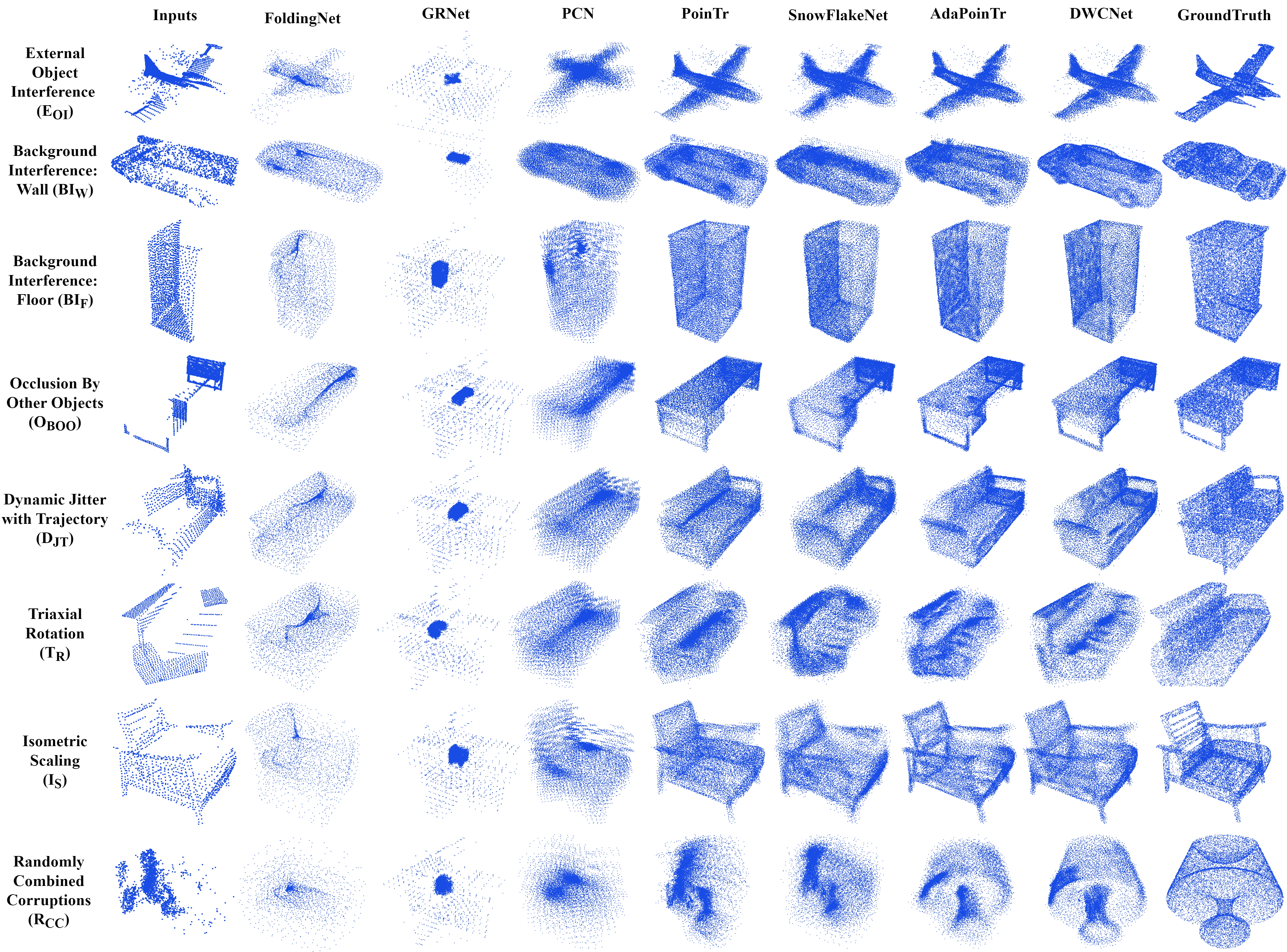}}
\caption{Example of results on CPCCD dataset after fine-tuning: multiple categories}  
\label{afafter}
\vspace{-0.1 in}
\end{figure*}

The $CD_{L1}$, $Fscore$ and $fidelity$ results after fine-tuning are shown in Tables \ref{resulttable2}, \ref{resulttable2fscore} and \ref{resulttable2fidelity} respectively. The $CD_{L2}$ results can be found in table \ref{resulttable2cdl2}.

\begin{table}[h!]
    \caption{Completion Results on CPCCD datasets after fine-tuning. The result shows $CD_{L1}$ results of the completion. The lower the $CD_{L1}$ the better the result. The superscripts 'PB', 'GB', 'CB' and 'TB' stand for Point-Based, Graph-Based, Convolution-Based and Transformer-Based respectively.} 
    \label{resulttable2} 
    \begin{scriptsize}
    \centering
    \setlength{\tabcolsep}{1.4pt}
    {\small
\resizebox{\linewidth}{!}{%
    {\begin{tabular}{c c c c c c c c c c c }
        \toprule
        Network ($CD_{L1}\downarrow$) & clean & $E_{OI}$ & $ {BI_W} $ & $ BI_F $ & $ O_{BOO} $ & $D_{JT}$ & $T_R$ & $I_S$ & $R_{CC}$ \\
        \midrule
        FoldingNet\textsuperscript{GB} & 16.228 & 16.912 & 16.828 & 16.672 & 16.304 & 16.359 & 16.288 & 17.461 & 19.584 \\
        PCN\textsuperscript{PB} & 31.932 & 32.683 & 34.090 & 32.872 & 31.973 & 32.044 & 32.098 & 33.225 & 36.114 \\
        GRNet\textsuperscript{TB} & 32.957 & 32.877 & 39.338 & 36.745 & 33.311 & 32.556 & 33.458 & 34.464 & 40.478 \\
        PoinTr\textsuperscript{TB} & 8.760 & 9.002 & 14.181 & 11.192 & 8.857 & 9.024 & 9.766 & 11.217 & 17.692 \\
        SnowFlakeNet\textsuperscript{TB} & 10.115 & 10.658 & 14.291 & 12.408 & 10.194 & 10.628 & 12.129 & 13.570 & 18.508 \\
        AdaPoinTr\textsuperscript{TB} & 8.150 & 8.309 & 8.724 & 8.568 & 8.211 & 8.584 & 8.918 & 9.196 & 10.317 \\
        DWCNet \textsuperscript{TB} (Ours) & \textbf{7.727} & \textbf{7.856} & \textbf{8.130} & \textbf{8.052} & \textbf{7.793} & \textbf{8.059} & \textbf{8.435} & \textbf{8.709} & \textbf{9.754} \\ 
        \bottomrule
          
    \end{tabular}    
    }}}
    \end{scriptsize}
\vspace{-0.15 in}
\end{table}

\begin{table}[h!]
    \caption{Completion Results on CPCCD datasets after fine-tuning. The result shows $Fscore$ results of the completion. The higher the $Fscore$ the better the result. The superscripts 'PB', 'GB', 'CB' and 'TB' stand for Point-Based, Graph-Based, Convolution-Based and Transformer-Based respectively.} 
    \label{resulttable2fscore} 
    \begin{scriptsize}
    \centering
    \setlength{\tabcolsep}{1.4pt}
    {\small
\resizebox{\linewidth}{!}{%
    {\begin{tabular}{c c c c c c c c c c c }
        \toprule
        Network ($fscore\uparrow$) & clean & $E_{OI}$ & $ {BI_W} $ & $ BI_F $ & $ O_{BOO} $ & $D_{JT}$ & $T_R$ & $I_S$ & $R_{CC}$ \\
        \midrule
        FoldingNet\textsuperscript{GB} & 0.279 & 0.256 & 0.273 & 0.268 & 0.276 & 0.276 & 0.276 & 0.257 & 0.219 \\
        PCN\textsuperscript{PB} & 0.182 & 0.170 & 0.174 & 0.169 & 0.182 & 0.178 & 0.175 & 0.163 & 0.142 \\
        GRNet\textsuperscript{TB} & 0.179 & 0.178 & 0.123 & 0.132 & 0.171 & 0.177 & 0.165 & 0.153 & 0.113 \\
        PoinTr\textsuperscript{TB} & 0.712 & 0.708 & 0.611 & 0.644 & 0.706 & 0.705 & 0.660 & 0.649 & 0.525 \\
        SnowFlakeNet\textsuperscript{TB} & 0.659 & 0.645 & 0.568 & 0.593 & 0.655 & 0.634 & 0.552 & 0.536 & 0.441 \\
        AdaPoinTr\textsuperscript{TB} & 0.756 & 0.750 & 0.733 & 0.738 & 0.751 & 0.734 & 0.710 & 0.694 & 0.646 \\
        DWCNet\textsuperscript{TB} (Ours) & \textbf{0.777} & \textbf{0.774} & \textbf{0.762} & \textbf{0.765} & \textbf{0.774} & \textbf{0.760} & \textbf{0.740} & \textbf{0.724} & \textbf{0.676} \\  
        \bottomrule
          
    \end{tabular}    
    }}}
    \end{scriptsize}
\vspace{-0.15 in}
\end{table}

\begin{table}[h!]
    \caption{Completion Results on CPCCD datasets after fine-tuning. The result shows $fidelity$ results of the completion. The lower the $fidelity$ the better the result. The superscripts 'PB', 'GB', 'CB' and 'TB' stand for Point-Based, Graph-Based, Convolution-Based and Transformer-Based respectively.} 
    \label{resulttable2fidelity} 
    \begin{scriptsize}
    \centering
    \setlength{\tabcolsep}{1.4pt}
    {\small
\resizebox{\linewidth}{!}{%
    {\begin{tabular}{c c c c c c c c c c c }
        \toprule
        Network ($fidelity\downarrow$) & clean & $E_{OI}$ & $ {BI_W} $ & $ BI_F $ & $ O_{BOO} $ & $D_{JT}$ & $T_R$ & $I_S$ & $R_{CC}$ \\
        \midrule
        FoldingNet\textsuperscript{GB} & 0.425 & 0.524 & 0.531 & 0.463 & 0.428 & 0.457 & 0.430 & 0.537 & 0.927 \\
        PCN\textsuperscript{PB} & 7.728 & 7.766 & 8.239 & 7.948 & 7.734 & 7.778 & 7.608 & 7.907 & 8.395 \\
        GRNet\textsuperscript{TB} & 9.208 & 9.078 & 10.212 & 9.562 & 9.310 & 8.951 & 9.071 & 8.936 & 10.128 \\
        PoinTr\textsuperscript{TB} & 0.190 & 0.237 & 1.549 & 0.547 & 0.193 & 0.214 & 0.235 & 0.799 & 2.924 \\
        SnowFlakeNet\textsuperscript{TB} & 0.308 & 0.393 & 1.130 & 0.604 & 0.308 & 0.331 & 0.401 & 0.804 & 1.971 \\
        AdaPoinTr\textsuperscript{TB} & 0.137 & 0.149 & 0.175 & 0.164 & 0.137 & 0.164 & 0.163 & 0.183 & 0.260 \\
        DWCNet\textsuperscript{TB} (Ours) & \textbf{0.127} & \textbf{0.135} & \textbf{0.144} & \textbf{0.148} & \textbf{0.128} & \textbf{0.142} & \textbf{0.147} & \textbf{0.165} & \textbf{0.238} \\  
        \bottomrule
          
    \end{tabular}    
    }}}
    \end{scriptsize}
\vspace{-0.15 in}
\end{table}

After fine-tuning on CPCCD, all networks adapted to the corruptions (Table~\ref{resulttable2}). Transformer-based networks outperformed point-based and convolution-based networks, adapting more quickly to the corruptions. Interestingly, after fine-tuning on randomly combined corruptions ($R_{CC}$), all algorithms exhibited similar performance across the CPCCD categories. Our method DWCNet with the noise management module performs really well on the majority of the corruptions. While the networks learned to tolerate the corruptions, they were unable to effectively remove them as can be seen in Figures \ref{chairfigsafter} and \ref{afafter}. 

\subsection{Evaluation on ScanObjectNN}

\begin{figure*}[t!]
\centerline{\includegraphics[width = \linewidth]{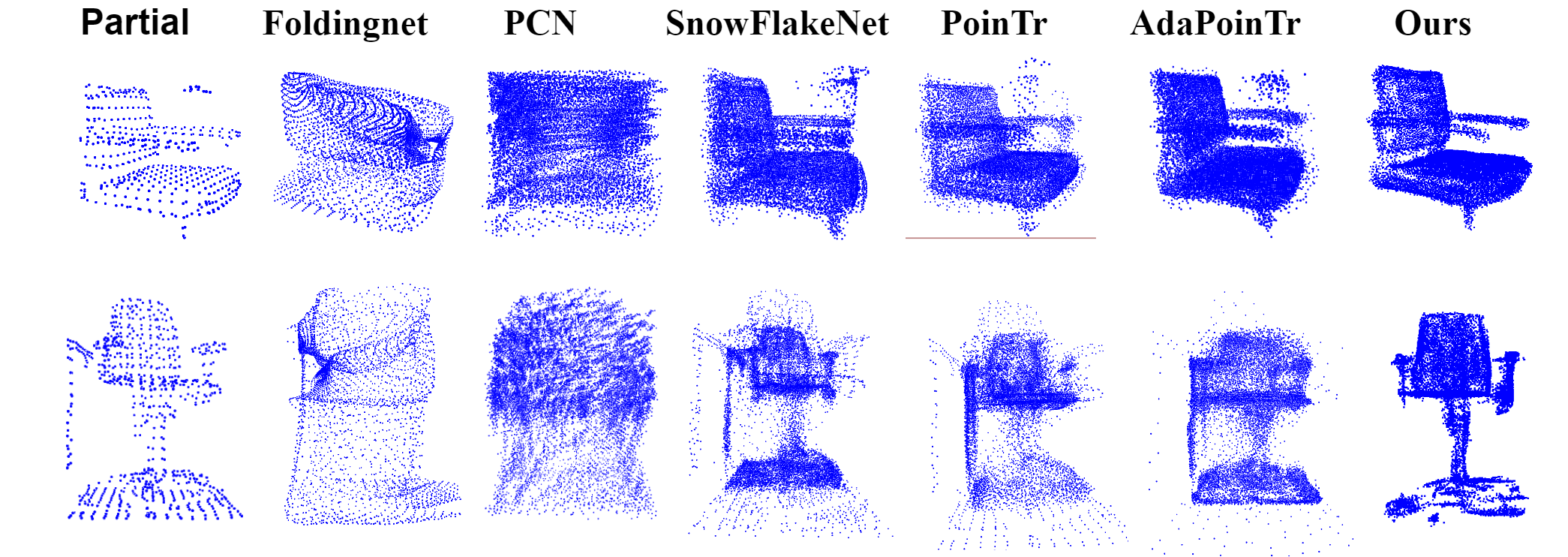}}
\caption{Some examples of results on ScanObjectNN dataset}
\label{scantestfig}
\vspace{-0.1 in}
\end{figure*}

Our goal is to enhance the robustness of SOTA completion algorithms to better handle noise and corruption in real-world point clouds. To evaluate this, we tested the performance on the real-world dataset ScanObjectNN \cite{scanobjectnn}.

We conducted experiments on 20 chairs, 20 tables, 20 desks, and 20 sofas from the ScanObjectNN dataset. Since real-world scans lack ground truth data, we present qualitative results in Figure~\ref{scantestfig}. The first row shows completion results for a real chair scan with external object interference ($E_{OI}$) and occlusion by other objects ($O_{BOO}$). Most models mistakenly treat the external corruption as part of the object, leading to noisy results and failing to complete the chair's legs. In the second row, we show results for a highly corrupted chair scan with background interference from walls and floors ($BI_W$ and $BI_F$) and isometric scale ($I_S$) corruption. Similar to the results in Figure~\ref{chairfigs}, PCN and FoldingNet handled scale corruption better than Transformer-based models (SnowFlakeNet, PoinTr, and AdaPoinTr). While PCN and FoldingNet improved in capturing global details after fine-tuning, they struggled with local details. Transformer-based models captured better local details but could not adapt to scale corruption and failed to ignore dense external object interference ($E_{OI}$).

\section{Ablation Study}
\label{ablationsection}

In this section, we perform an ablation study on the baseline model, components, and loss functions of NMM. First, we present the results of the ablation study in Table~\ref{ablation_table}, comparing DWCNet with and without the Noise Management Module (NMM). On the PCN dataset, DWCNet without NMM (equivalent to AdapoinTr) achieves a Chamfer Distance of 8.351, while integrating NMM reduces it to 8.122. Though the improvement is modest, it clearly demonstrates the added value of NMM. The impact is even more pronounced on the CPCCD dataset, where the Chamfer Distance drops significantly from 13.403 to 10.263 after 200 epochs of training with NMM. These results highlight NMM’s effectiveness in handling data corruptions and improving point cloud completion, reinforcing its essential role in enhancing robustness and accuracy in real-world applications.

\begin{table}[h!]
\vspace{-0.1 in}
    \caption{Ablation Studies without (w/o) and with NMM}
    \label{ablation_table}
    \centering
    {\small{
    \resizebox{\linewidth}{!}{
    \begin{tabular}{p{0.3\linewidth}p{0.4\linewidth}p{0.15\linewidth}}
    \toprule
    Dataset($CD_{L1}\downarrow)$ & \centering DWCNET w/o NMM  & DWCNET \\
    \midrule
    PCN & \centering 8.351 & 8.122 \\
    $Rcc$& \centering 13.403 & 10.263\\
    \bottomrule
    \end{tabular}
    }}}
\vspace{-0.15 in}
\end{table}

\subsection{Path Modeling}
In this section, we discuss the effect of having only a clean path, only noisy path or both paths in the architecture. For the noisy path experiment, we had to calculate the clean features, $f_{clean}$ that are normally produced by the clean path, by subtracting the noisy features $f_{noisy}$ from the input features $f_{i}$ produced by the geometric-aware transformer instead. 

\begin{equation}
    f_{\text{clean}} = {f}_{\text{i}} - {f}_{\text{noisy}}
\end{equation}

\begin{table}[h!]
    \caption{Ablation studies on the baseline model}
    \label{ablation_table2}
    \centering
    {\small{
    \begin{tabular}{p{0.3\linewidth}p{0.1\linewidth}p{0.15\linewidth}p{0.18\linewidth}}
    \toprule
    Path  & \centering $CD_{L1}\downarrow$ & $Fscore\uparrow$ & $Fidelity \downarrow$ \\
    \midrule
    No NMM &  13.403 & 0.56 & \\
    Clean path only & 10.255 & 0.646 & 0.259 \\
    Noisy path only & 28.777 & 0.398 & 7.579\\
    DWCNet & 10.263 & 0.646 & 0.268\\
    \bottomrule
    \end{tabular}
    }}
\vspace{-0.15 in}
\end{table}

All experiments were conducted under the same conditions, with a temperature scaling value of t = 0.1 and a training duration for 200 epochs. We observed that the clean path contributes the most to the training, and the worst results were observed from the noisy path. The difference between the clean path only and the using both paths was not too significant, so we performed the ablation study below on the architecture components and contrastive loss for further investigation into the effectiveness of each path.

\begin{table*}[th!]
\caption{Ablation studies on the effect of the temperature scaling parameter $t$ at 100 epochs}
\centering
\label{ablation_table4}
\resizebox{\linewidth}{!}{
    \begin{tabular}{lccccccccccccc}
    \toprule
    \textbf{$t$ (100 epochs)} & 0.0000005 & 0.05 & 0.1 & 0.25 & 0.35 & 0.5 & 0.75 & 0.95 & 1.0 & 10 & 100 & 1000 & 10000\\
    \midrule
    Rcc (CD) $\downarrow$ & 31.608 & 11.773 & 11.503 & 11.610 & 11.502 & 11.432 & 11.670 & 11.592 & 11.438 & 11.279 & 11.2 & 11.528 & 11.712 \\
    Rcc (Fscore) $\uparrow$ & 0.334 & 0.570 & 0.586 & 0.584 & 0.586 & 0.582 & 0.581 & 0.583 & 0.592 & 0.594 & 0.598 & 0.584 & 0.573 \\
    \bottomrule
\end{tabular}
}
\vspace{-0.15in}
\end{table*}

\subsection{Architecture Components}

To evaluate the contributions of attention mechanisms and multiscale convolutions in our model, we conducted a series of ablation experiments. In the ``No Attention" configuration, we replaced the transformer module in the clean path with a simple MLP layer and trained the model for 100 epochs. In the ``Single-Scale Convolution (SSC)" experiment, we substituted the multiscale convolution (MSC) in the noisy path with a standard single-scale convolution, also training for 100 epochs. Finally, in the ``No Attention + SSC" setup, we combined both modifications—using the MLP in the clean path and the SSC in the noisy path. All experiments were conducted under the same conditions, with a temperature scaling value of t = 0.1 and a training duration of 100 epochs. 

\begin{table}[h!]
    \caption{Ablation studies on architecture components. SSC refers to Single Scale Convolution and MSC stands for Multi Scale Convolution}
    \label{ablation_table3}
    \centering
    \resizebox{\linewidth}{!}{
    {\small{
    \begin{tabular}{p{0.4\linewidth}p{0.2\linewidth}p{0.2\linewidth}}
    \toprule
    Path  & \centering $CD_{L1}\downarrow$ & $Fscore\uparrow$\\
    \midrule
    No attention & 20.846 & 0.357  \\
    SSC & 11.503 & 0.589 \\
    No attention and SSC  & 11.47 & 0.585 \\
    DWCNet & 11.503 & 0.586 \\
    \bottomrule
    \end{tabular}
    }}}
\vspace{-0.15in}
\end{table}

We found that while having no attention affected NMM's performance negatively, the use of Single Scale convolution or Multiscale convolution did not have any significant effect. This finding supports our observation that the clean path contributes the most in early training. 

\subsection{Contrastive Loss Variants}

From the ablation studies, we observed that the \textit{clean path}, which contributes to the \textit{Positive Loss}, plays a crucial role in completing highly corrupted partial point clouds. This effectiveness is likely due to the availability of clean partial inputs that provide strong supervision. The Positive Loss encourages alignment between clean features and their corresponding ground truth, reinforcing meaningful representations.

To complement this, we introduced the \textit{Negative Loss} to penalize alignment between mismatched (clean-noisy) pairs and promote discriminative feature learning. Although the Positive Loss proved more impactful, removing the Negative Loss risks \textit{representation collapse}, where features converge to trivial, non-informative embeddings due to the absence of contrastive pressure \cite{collapse}.

To better balance these losses, we investigated the effect of the \textit{temperature scaling parameter} $t$, which modulates the sharpness of the Negative Loss (see Table~\ref{ablation_table4}). At extremely low values (e.g., $t = 0.0000005$), the Negative Loss became excessively large, leading to poor completion performance. While differences between $t = 0.05$ and $t = 1.0$ were subtle over 100 epochs, a trend toward improved performance with higher $t$ values emerged. We extended training to 350 epochs and evaluated $t = 0.1$, $t = 0.5$, $t = 1.0$ and $t = 100.0$ (see Table~\ref{ablation_table5}). We found that \textbf{$t = 1.0$} and \textbf{$t = 100.0$} yielded the most favorable results on the $Rcc$ category, with chamfer distances of 9.754 and 9.752, respectively. While \textbf{$t = 100.0$} offered a marginal advantage on $Rcc$, the model trained at \textbf{$t = 1.0$} outperformed across all other CPCCD categories and was therefore selected as the optimal configuration as shown in Table \ref{resulttablet1vst100}.

\begin{table}[h!]
\caption{Ablation studies on the effect of the temperature scaling parameter $t$ at 350 epochs}
\centering
\label{ablation_table5}
\resizebox{\linewidth}{!}{
\begin{tabular}{lccccc}
\toprule
\textbf{$t(350 epochs)$} & 0.1 & 0.5 & 1.0 & 100 & clean loss\\
\midrule
Rcc (CD$\downarrow$) & 10.328 &  9.957 & 9.754 & 9.752 & 9.989\\
Rcc (Fscore$\uparrow$) & 0.642 & 0.663 & 0.676 & 0.676 & 0.661 \\
\bottomrule
\end{tabular}
}
\vspace{-0.15in}
\end{table}

\begin{table}[h!]
    \caption{Completion results for DWCNet ($t=1$) and DWCNet ($t=100$) at 350 epochs} 
    \label{resulttablet1vst100} 
    \begin{scriptsize}
    \centering
    \setlength{\tabcolsep}{1.4pt}
    {\small
\resizebox{\linewidth}{!}{%
    {\begin{tabular}{c c c c c c c c c c c }
        \toprule 
        Network (CDL1$\downarrow$) & clean & $E_{OI}$ & $ {BI_W} $ & $ BI_F $ & $ O_{BOO} $ & $D_{JT}$ & $T_R$ & $I_S$ & $R_{CC}$ \\
        \midrule
        DWCNet(t=100) \textsuperscript{TB} & 7.756 & 7.921 & 8.193 & 8.15 & 7.839 & 8.126 & 8.499 & 8.791 & \textbf{9.752} \\
        DWCNet(t=1) \textsuperscript{TB} & \textbf{7.727} & \textbf{7.856} & \textbf{8.130} & \textbf{8.052} & \textbf{7.793} & \textbf{8.059} & \textbf{8.435} & \textbf{8.709} & 9.754 \\ 
        \hline 
    \end{tabular}    
    }}}
    \end{scriptsize}
\vspace{-0.15 in}
\end{table}

\section{Conclusion and Future Work}

In this paper, we introduce the Corrupted Point Cloud Completion Dataset (CPCCD) -- the first dataset specifically curated to evaluate the robustness of point cloud completion algorithms under noise-induced corruption. Through comprehensive experiments, we demonstrate that fine-tuning existing completion models on CPCCD substantially improves their resilience to noise commonly encountered in real-world scenarios. To further advance robustness, we propose a novel Noise Management Module (NMM), which facilitates joint denoising and completion of severely corrupted partial point clouds within our robust completion framework, DWCNet. While our approach yields promising results, we recognize the need for more accurate simulation of real-world noise characteristics and structural degradation. As part of future work, we plan to expand CPCCD by incorporating additional object categories from established clean completion benchmarks such as ShapeNet34 and ShapeNet55. Furthermore, we aim to enhance the noise-removal and structure-preservation capabilities of completion algorithms. By addressing these challenges, we seek to underscore the practical significance of completing highly corrupted partial point clouds -- an area that remains underexplored in the current literature.




\section*{Acknowledgments}
This work was funded by EPSRC grant number EP/S021892/1 and Beam (Previously Vaarst) (\href{http://www.beam.global}{beam.global}). Tam is partially supported by a Royal Society fund IEC/NSFC/211159. 
For the purpose of open access the authors have applied a Creative Commons Attribution (CC BY) licence to any Author Accepted Manuscript version arising from this submission. 

\bibliographystyle{elsarticle-num-names}
\bibliography{main}

\vspace{-0.1 in}
\appendix
\label{appendixmain}

\section{Quantitative Results for the Robustness Test}
\label{appendix2}

\subsection{Before Fine-tuning}

In this section we present $CD_{L2}$ results for the robustness evaluation of the completion networks before fine-tuning in Table~\ref{resulttable1cdl2}. We find that the results are consistent with those in Table~\ref{resulttable1}. 

\begin{table}[h!]
    \caption{Completion Results on CPCCD datasets. The result shows $CD_{L2}$ results of the completion. The lower the $CD_{L2}$ the better the result. The superscripts 'PB', 'GB', 'CB' and 'TB' stand for Point-Based, Graph-Based, Convolution-Based and Transformer-Based respectively.} 
    \label{resulttable1cdl2} 
    \begin{scriptsize}
    \centering
    \setlength{\tabcolsep}{1.4pt}
    {\small
\resizebox{\linewidth}{!}{%
    {\begin{tabular}{c c c c c c c c c c c }
        \toprule 
        Network ($CD_{L2}\downarrow$) & clean & $E_{OI}$ & $ {BI_W} $ & $ BI_F $ & $ O_{BOO} $ & $D_{JT}$ & $T_R$ & $I_S$ & $R_{CC}$ \\
        \midrule
        FoldingNet\textsuperscript{GB} & 6.78 & 7.096 & 7.096 & 11.885 & 6.765 & 6.086 & 6.81 & 9.466 & 12.048 \\
        PCN\textsuperscript{PB} & 10.5 & 10.976 & 17.489 & 12.731 & 10.493 & 10.088 & 10.182 & 12.412 & 18.107 \\
        GRNet\textsuperscript{TB} & 5.009 & 5.318 & 15.288 & 8.447 & 5.146 & 4.854 & 5.357 & 11.2 & 23.732 \\
        PoinTr\textsuperscript{TB} & 0.336 & \textbf{1.053} & \textbf{5.219} & \textbf{1.615} & 0.338 & 0.54 & 0.835 & 3.496 & 7.757 \\
        SnowFlakeNet\textsuperscript{TB} & 0.312 & 1.479 & 6.042 & 2.799 & 0.318 & \textbf{0.502} & \textbf{0.786} & \textbf{2.508} & 7.376 \\
        AdaPoinTr\textsuperscript{TB} & \textbf{0.191} & 1.468 & 5.577 & 2.387 & 0.195 & 0.604 & 0.931 & 2.917 & \textbf{7.165} \\
        DWCNet\textsuperscript{TB} & 0.193 & 1.451 & 5.61 & 2.451 & \textbf{0.193} & 0.63 & 0.935 & 31673.3 & 112.697 \\
        \bottomrule    
        \hline 
    \end{tabular}    
    }}}
    \end{scriptsize}
\vspace{-0.15 in}
\end{table}

\subsection{After Fine-tuning}
In this section we present $CD_{L2}$ results for the robustness evaluation of the completion networks after fine-tuning on CPCCD. We find that the results are consistent with those in Table \ref{resulttable2}, \ref{resulttable2fscore} and \ref{resulttable2fidelity}.

\begin{table}[h!]
    \caption{Completion Results on CPCCD datasets after fine-tuning. The result shows $CD_{L2}$ results of the completion. The lower the $CD_{L2}$ the better the result. The superscripts 'PB', 'GB', 'CB' and 'TB' stand for Point-Based, Graph-Based, Convolution-Based and Transformer-Based respectively.} 
    \label{resulttable2cdl2} 
    \begin{scriptsize}
    \centering
    \setlength{\tabcolsep}{1.4pt}
    {\small
\resizebox{\linewidth}{!}{%
    {\begin{tabular}{c c c c c c c c c c c }
        \toprule
        Network ($CD_{L2}\downarrow$) & clean & $E_{OI}$ & $ {BI_W} $ & $ BI_F $ & $ O_{BOO} $ & $D_{JT}$ & $T_R$ & $I_S$ & $R_{CC}$ \\
        \midrule
        FoldingNet\textsuperscript{GB} & 0.901 & 1.002 & 1.024 & 0.958 & 0.905 & 0.927 & 0.912 & 1.058 & 1.492 \\
        PCN\textsuperscript{PB} & 7.977 & 8.033 & 8.541 & 8.221 & 7.985 & 8.031 & 7.862 & 8.205 & 8.745 \\
        GRNet\textsuperscript{TB} & 9.386 & 9.257 & 10.600 & 9.886 & 9.494 & 9.124 & 9.277 & 9.192 & 10.598 \\
        PoinTr\textsuperscript{TB} & 0.314 & 0.360 & 1.765 & 0.697 & 0.320 & 0.339 & 0.373 & 0.942 & 3.184 \\
        SnowFlakeNet\textsuperscript{TB} & 0.471 & 0.550 & 1.341 & 0.780 & 0.474 & 0.490 & 0.581 & 1.006 & 2.211 \\
        AdaPoinTr\textsuperscript{TB} & 0.272 & 0.283 & 0.328 & 0.306 & 0.273 & 0.309 & 0.308 & 0.328 & 0.443 \\
        DWCNet\textsuperscript{TB} (Ours) & \textbf{0.246} & \textbf{0.256} & \textbf{0.269} & \textbf{0.269} & \textbf{0.250} & \textbf{0.266} & \textbf{0.277} & \textbf{0.296} & \textbf{0.403} \\
        \bottomrule
                  
    \end{tabular}    
    }}}
    \end{scriptsize}
\vspace{-0.15 in}
\end{table}

\section{Qualitative Results for the Robustness Test}
\label{appendix3}
\begin{figure*}[th!]
\centerline{\includegraphics[width=\linewidth]{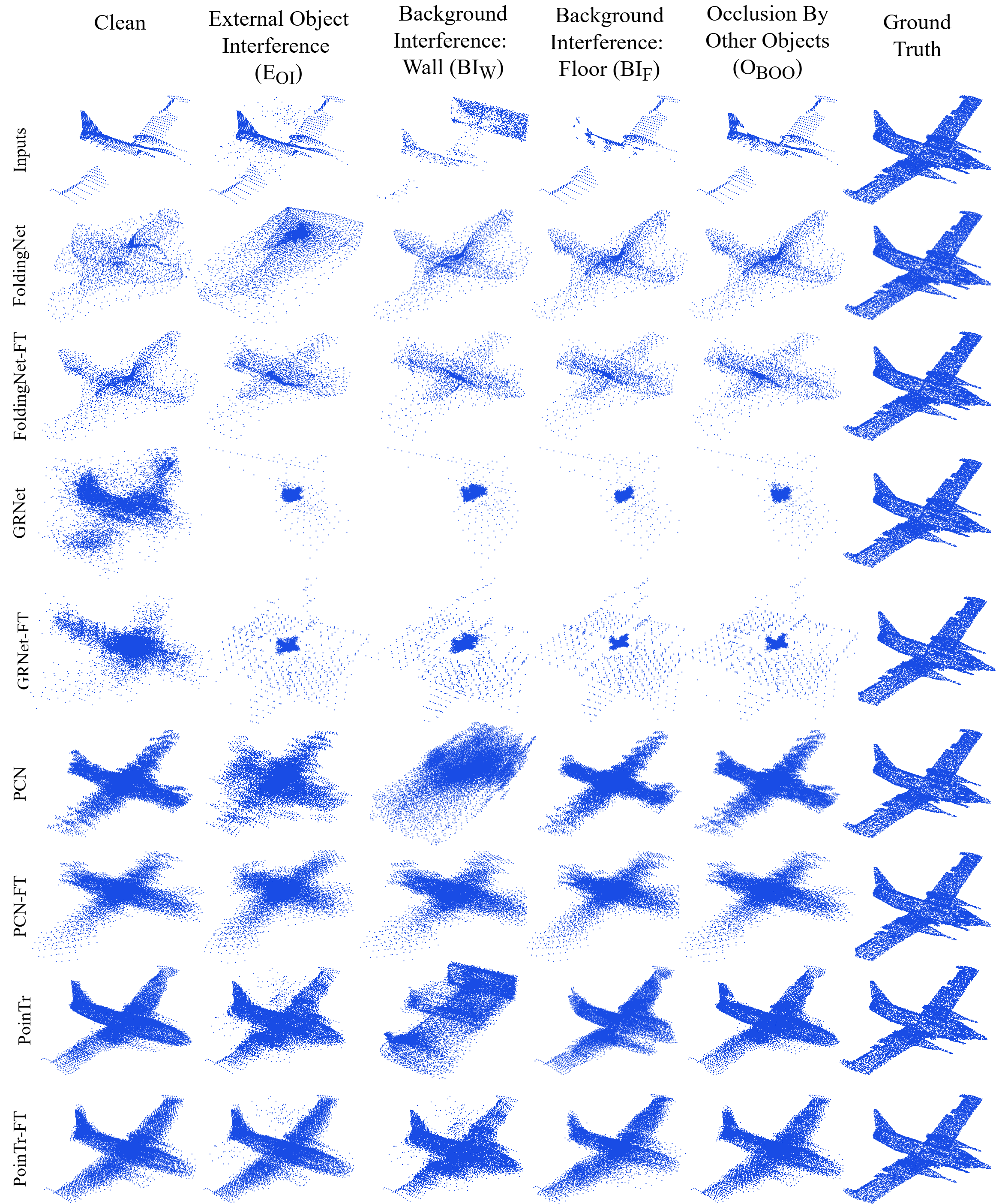}}
\caption{Qualitiative robustness evaluation result  examples (FoldingNet, GRNet, PCN, PoinTr): on an Aeroplane object on corruptions $E_{OI}$, $ {BI_W} $ , $ BI_F $ , $ O_{BOO} $ } 
\label{01dwcall}
\end{figure*}

\begin{figure*}[th!]
\centerline{\includegraphics[width=\linewidth]{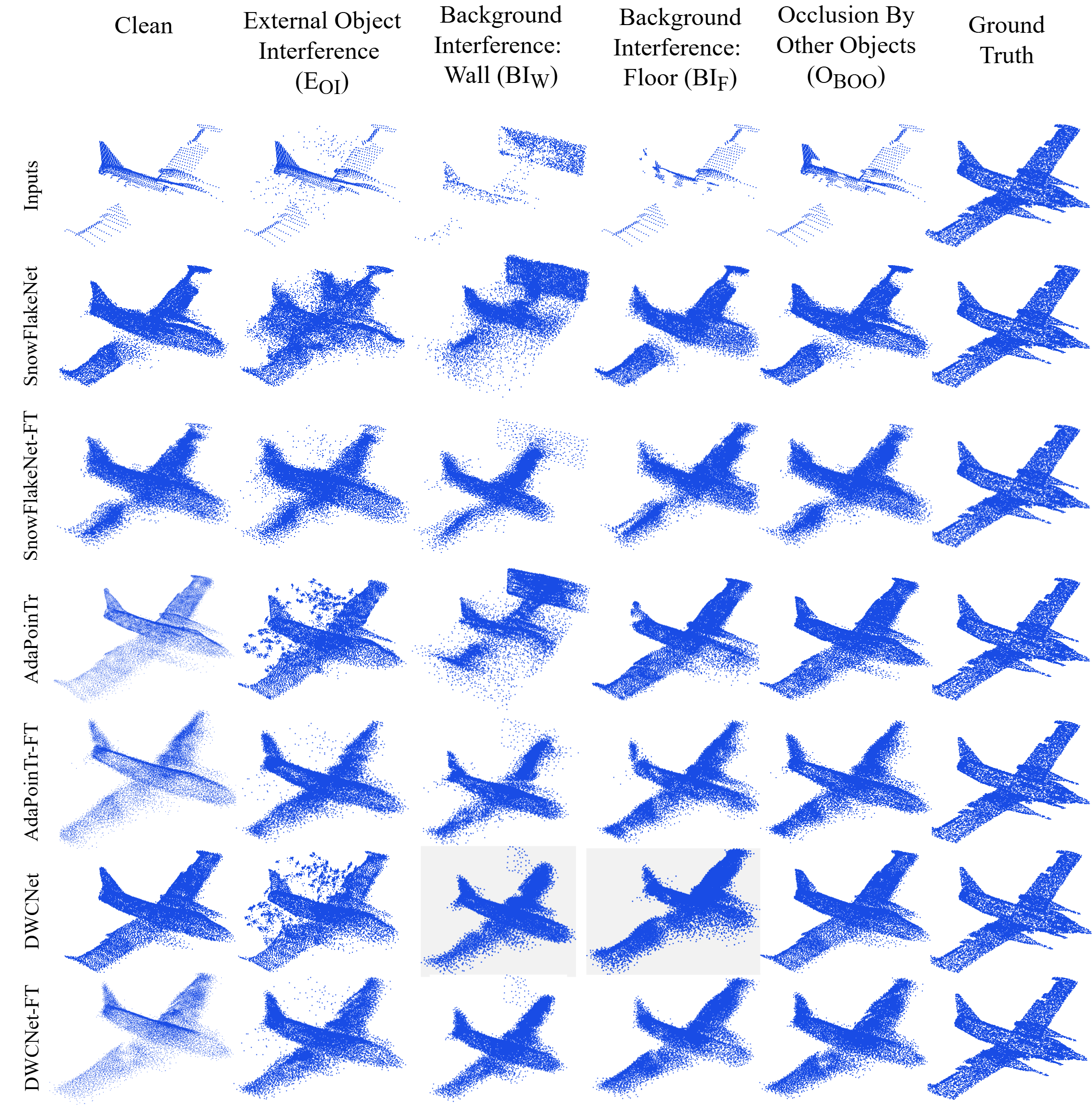}}
\caption{Qualitiative robustness evaluation result  examples (SnowFlakeNet, AdaPoinTr, DWCNet): on an Aeroplane object on corruptions $E_{OI}$, $ {BI_W} $ , $ BI_F $ , $ O_{BOO} $ } 
\label{02dwcall}
\vspace{-0.1in}
\end{figure*}

\begin{figure*}[th!]
\vspace{-0.1in}
\centerline{\includegraphics[width=\linewidth]{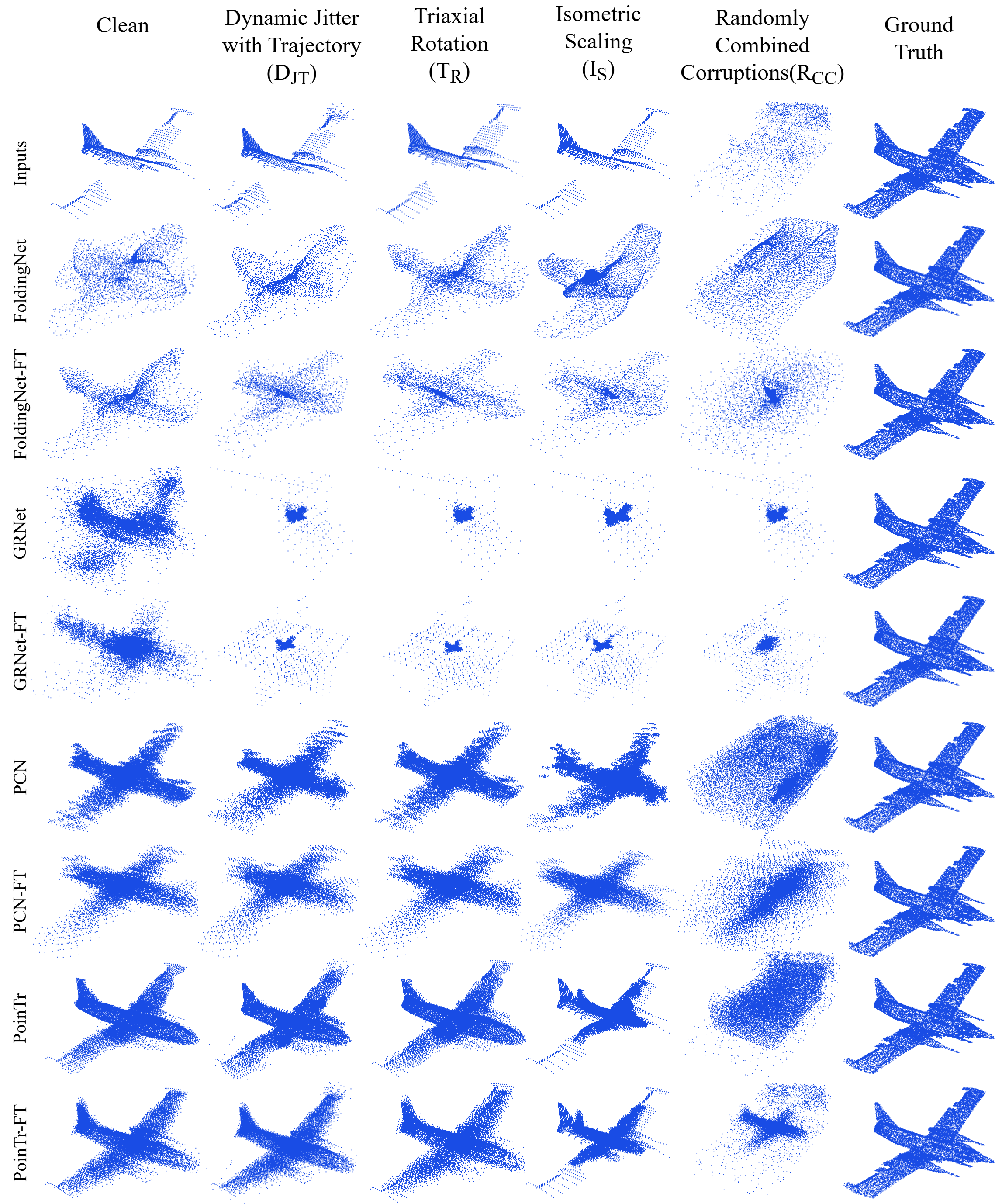}}
\caption{Qualitiative robustness evaluation result examples (FoldingNet, GRNet, PCN, PoinTr): on an Aeroplane object on corruptions $D_{JT}$, $T_R$, $I_S$, $R_{CC}$} 
\label{03dwcall}
\vspace{-0.1in}
\end{figure*}

\begin{figure*}[th!]
\vspace{-0.1in}
\centerline{\includegraphics[width=\linewidth]{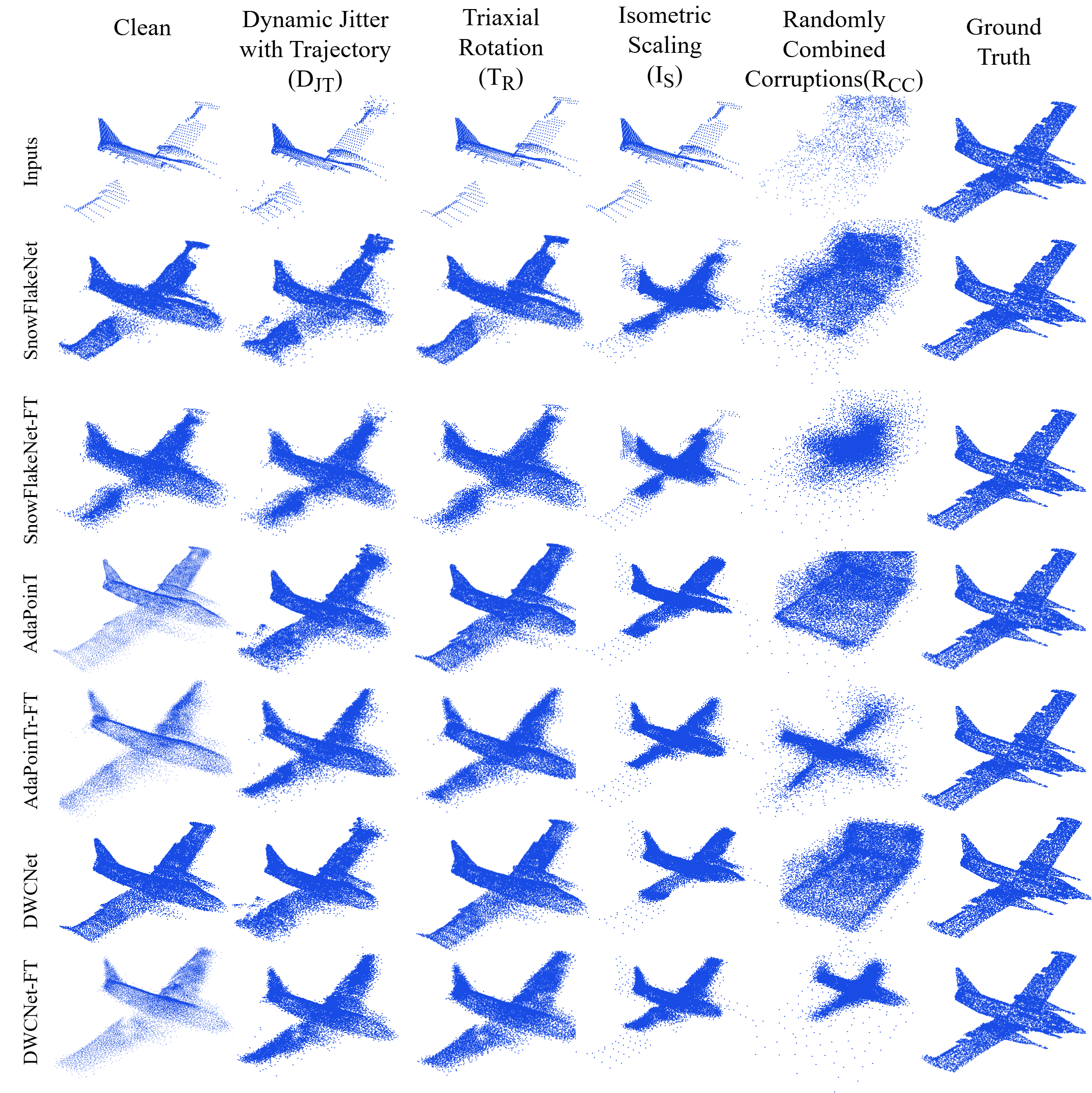}}
\caption{Qualitiative robustness evaluation result  examples (SnowFlakeNet, AdaPoinTr, DWCNet): on an Aeroplane object on corruptions $D_{JT}$, $T_R$, $I_S$, $R_{CC}$} 
\label{04dwcall}
\vspace{-0.1in}
\end{figure*}

\begin{figure*}[th!]
\vspace{-0.1in}
\centerline{\includegraphics[width=\linewidth]{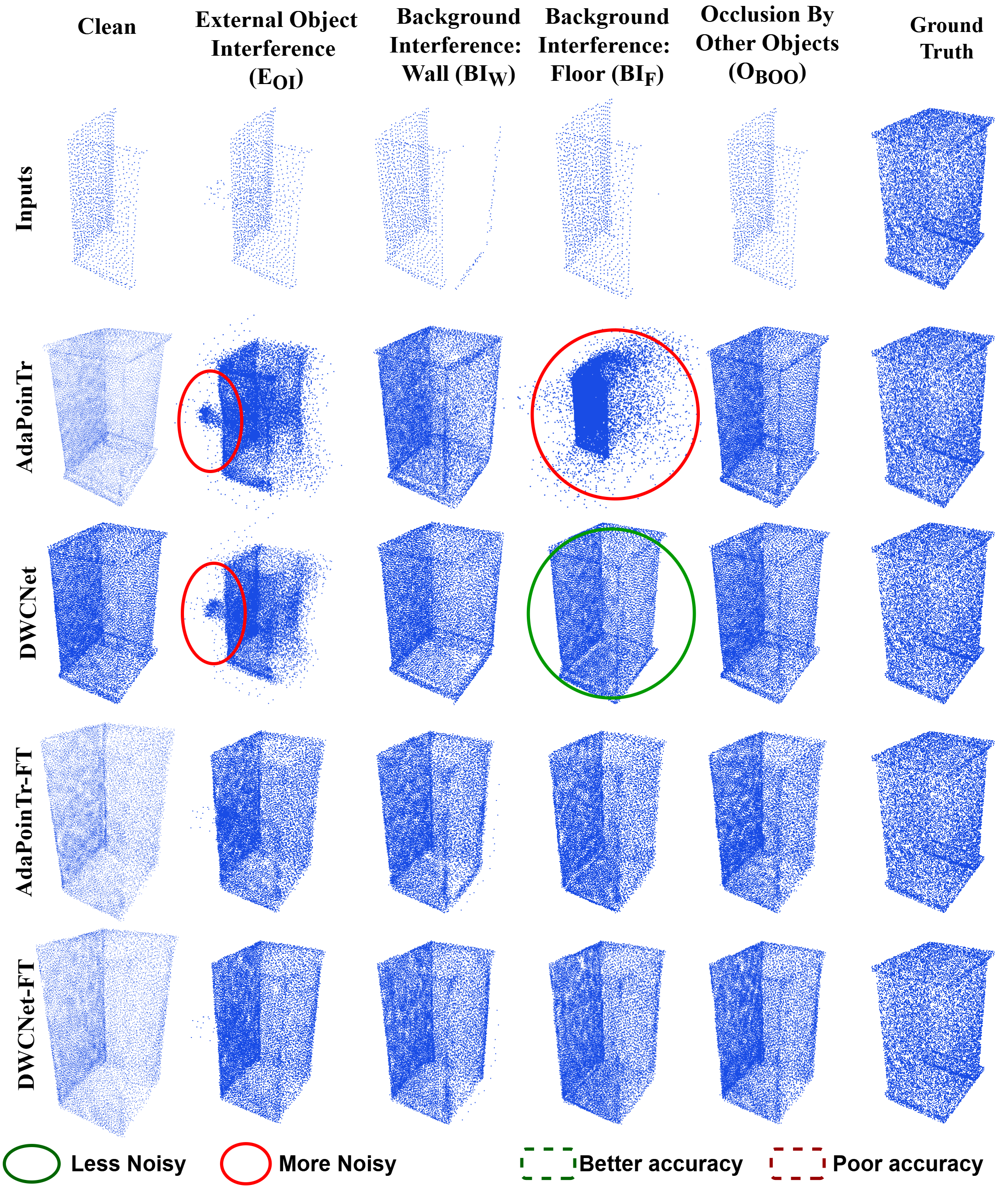}}
\caption{Comparison of AdaPoinTr and DWCNet: Cabinet category ($E_{OI}, BI_w , BI_F, O_{BOO}$)} 
\label{02dwca}
\vspace{-0.1in}
\end{figure*}

\begin{figure*}[th!]
\vspace{-0.1in}
\centerline{\includegraphics[width=\linewidth]{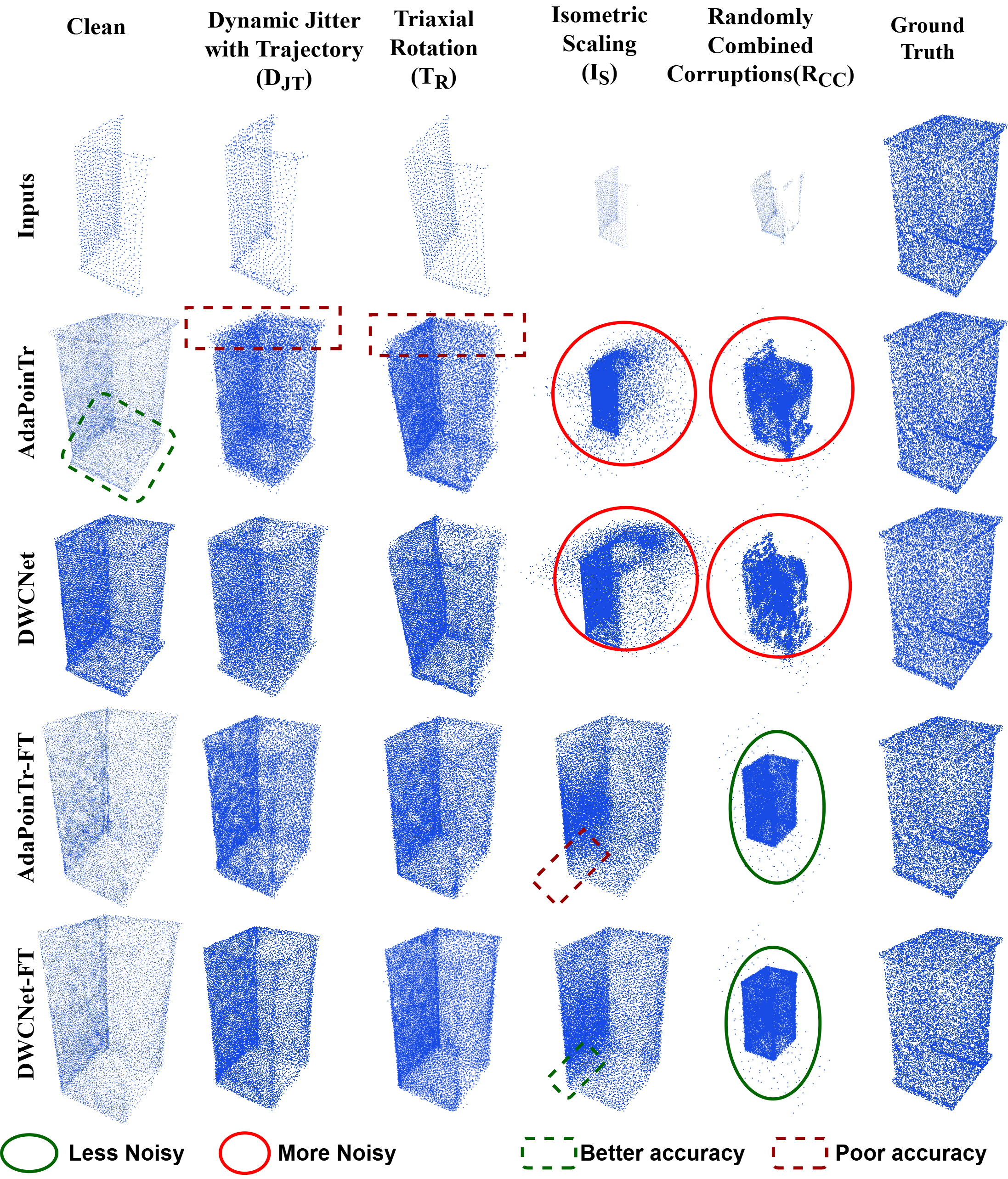}}
\caption{Comparison of AdaPoinTr and DWCNet: Cabinet category ($D_{JT}, T_R, I_S, R_{CC}$)} 
\label{02dwcb}
\vspace{-0.1in}
\end{figure*}

\begin{figure*}[th!]
\vspace{-0.1in}
\centerline{\includegraphics[width=\linewidth]{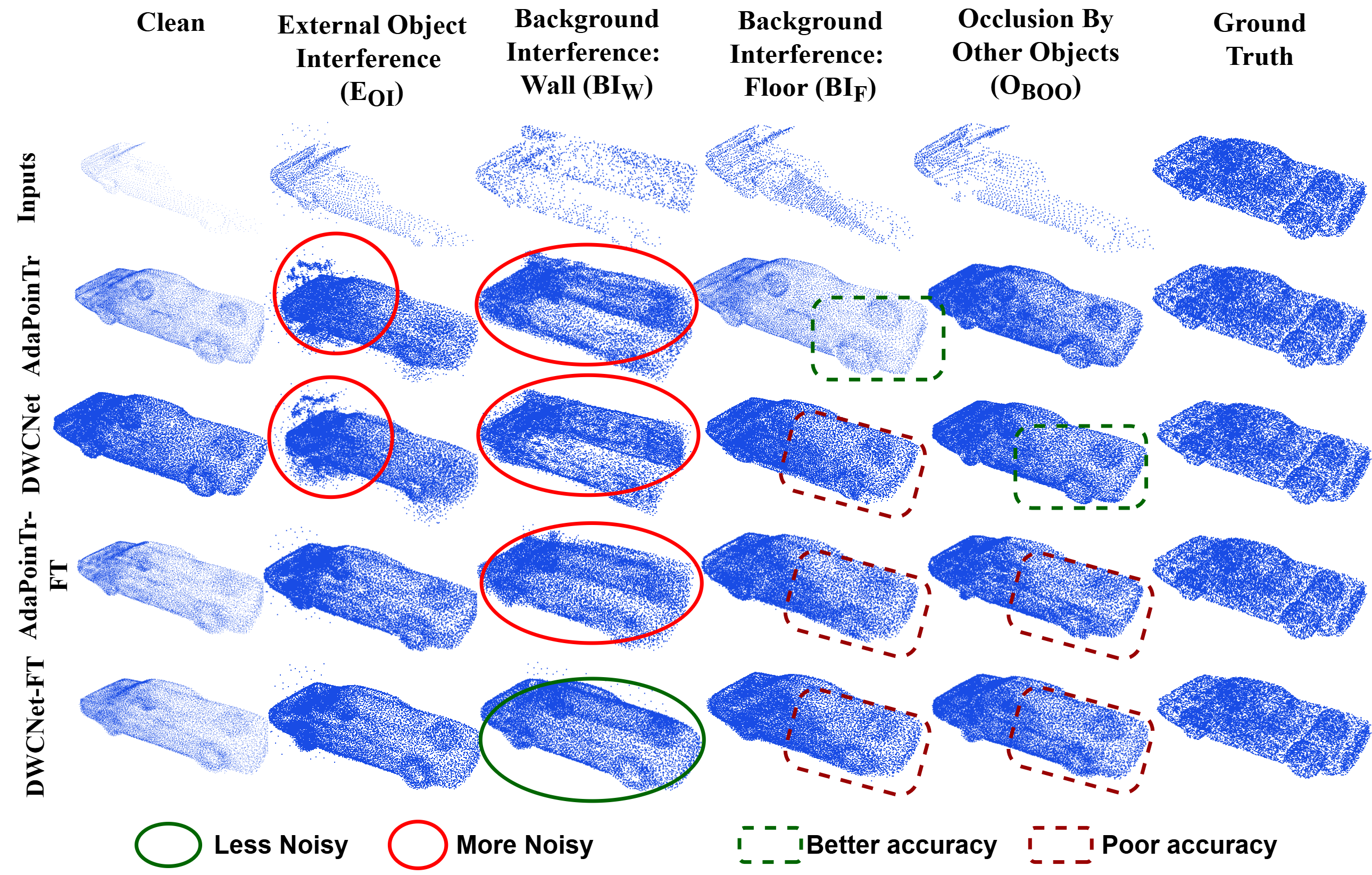}}
\caption{Comparison of AdaPoinTr and DWCNet: Car Category ($E_{OI}, BI_w , BI_F, O_{BOO}$)} 
\label{03dwca}
\vspace{-0.1in}
\end{figure*}

\begin{figure*}[th!]
\vspace{-0.1in}
\centerline{\includegraphics[width=\linewidth]{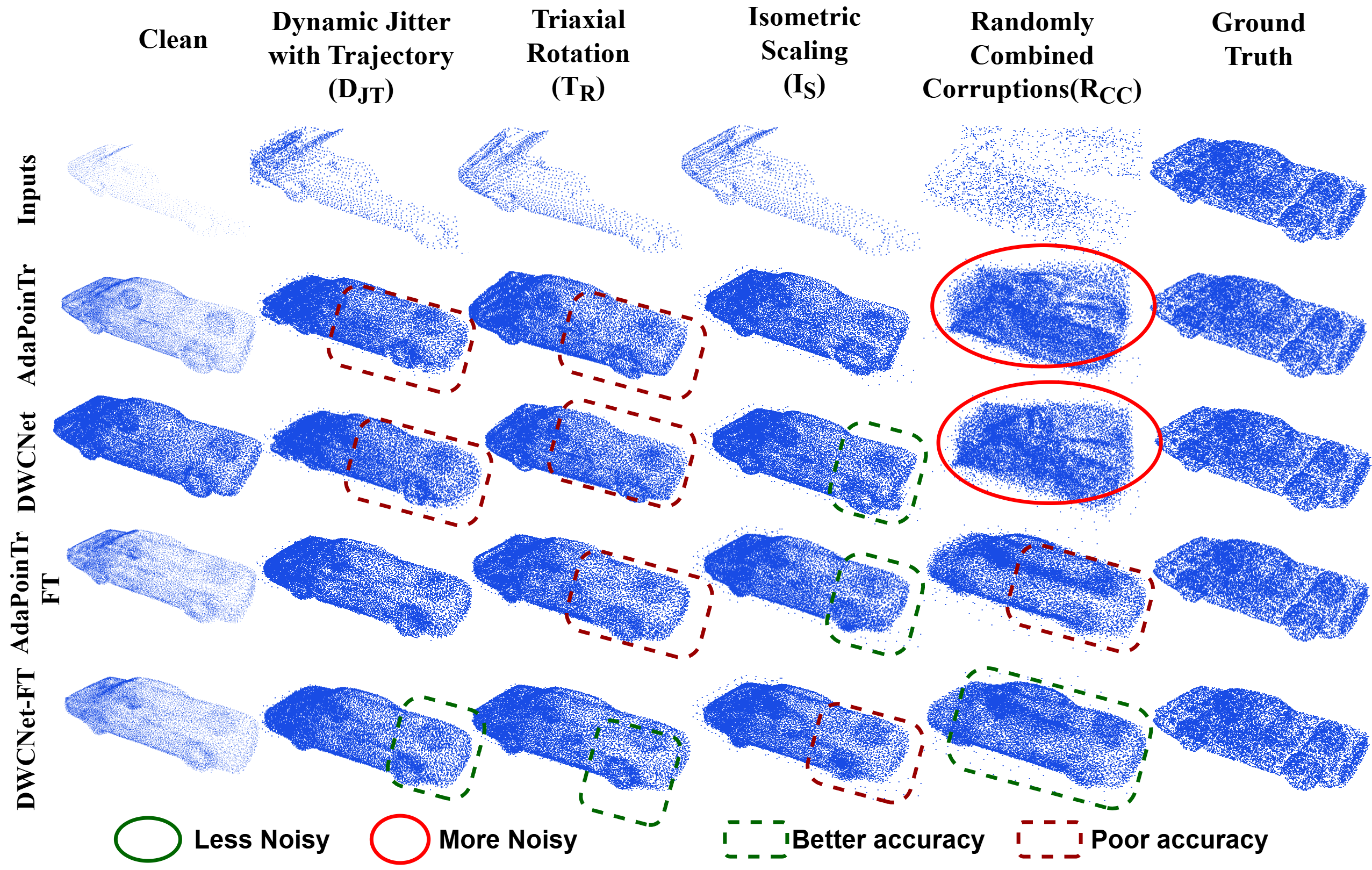}}
\caption{Comparison of AdaPoinTr and DWCNet: Car Category ($D_{JT}, T_R, I_S, R_{CC}$)} 
\label{03dwcb}
\vspace{-0.1in}
\end{figure*}

\begin{figure*}[th!]
\vspace{-0.1in}
\centerline{\includegraphics[width=\linewidth]{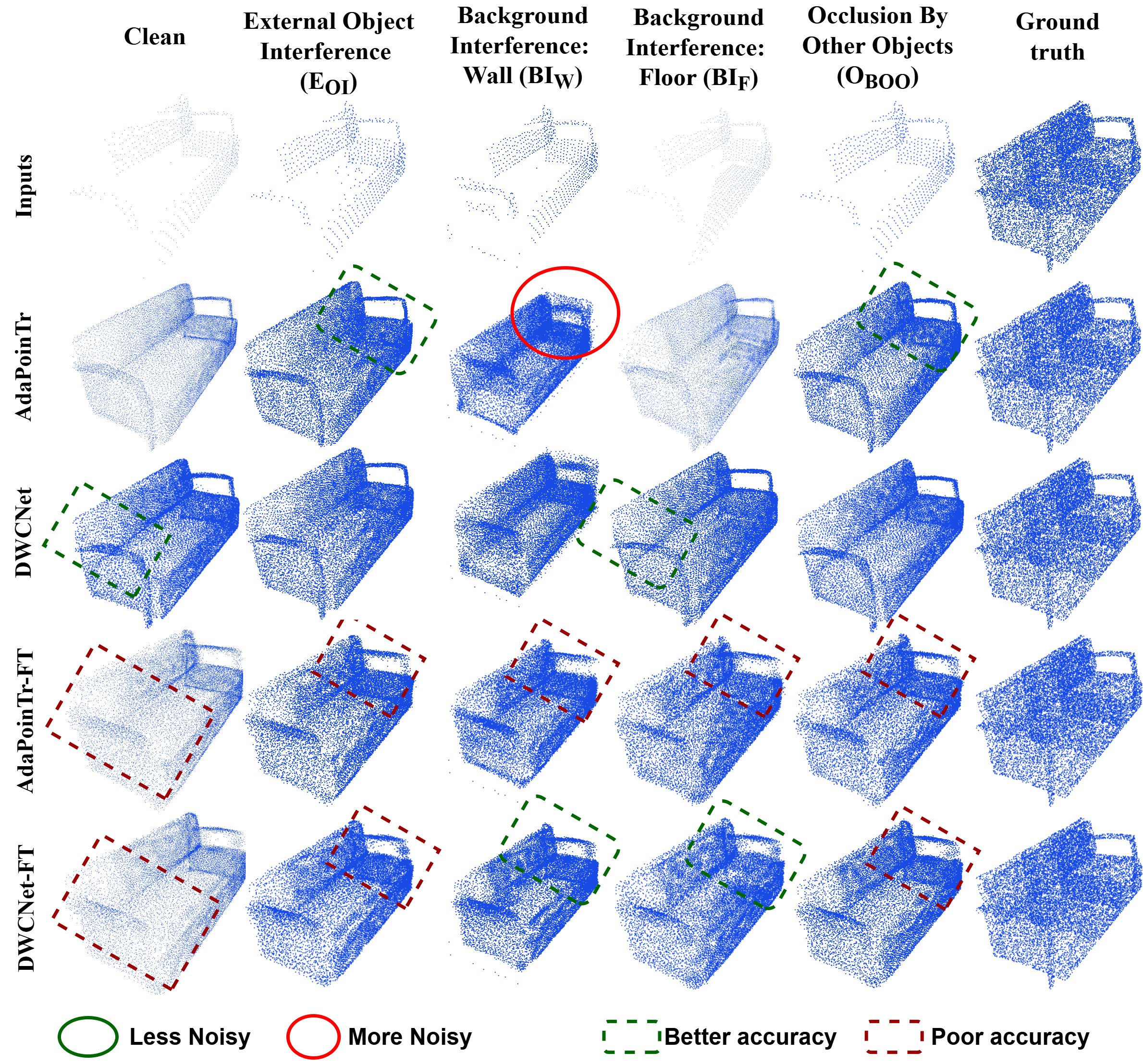}}
\caption{Comparison of AdaPoinTr and DWCNet: Chair Category ($E_{OI}, BI_w , BI_F, O_{BOO}$)} 
\label{04dwca}
\vspace{-0.1in}
\end{figure*}

\begin{figure*}[th!]
\vspace{-0.1in}
\centerline{\includegraphics[width=\linewidth]{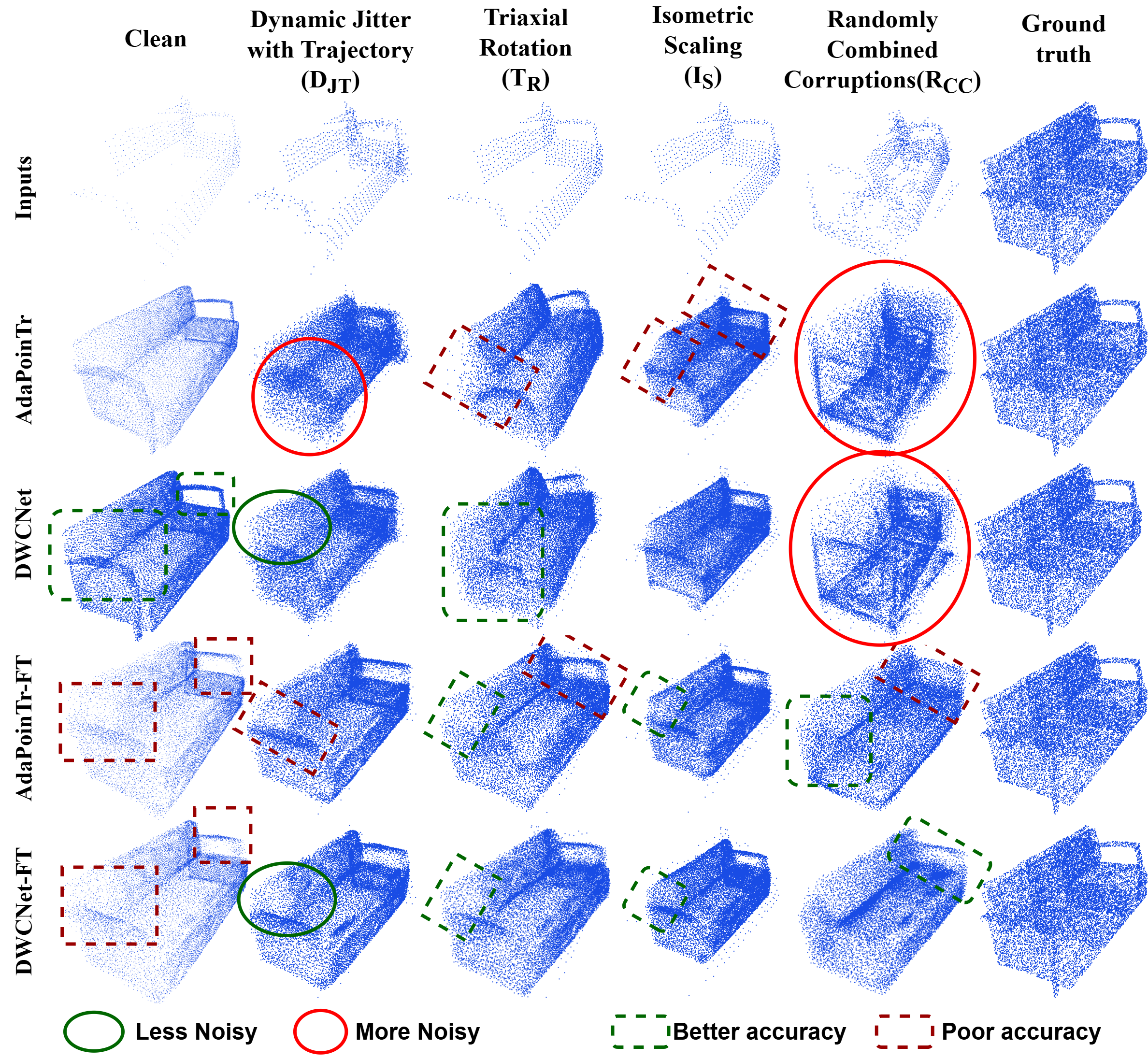}}
\caption{Comparison of AdaPoinTr and DWCNet: Chair Category ($D_{JT}, T_R, I_S, R_{CC}$)} 
\label{04dwcb}
\vspace{-0.1in}
\end{figure*}

\begin{figure*}[th!]
\vspace{-0.1in}
\centerline{\includegraphics[width=\linewidth]{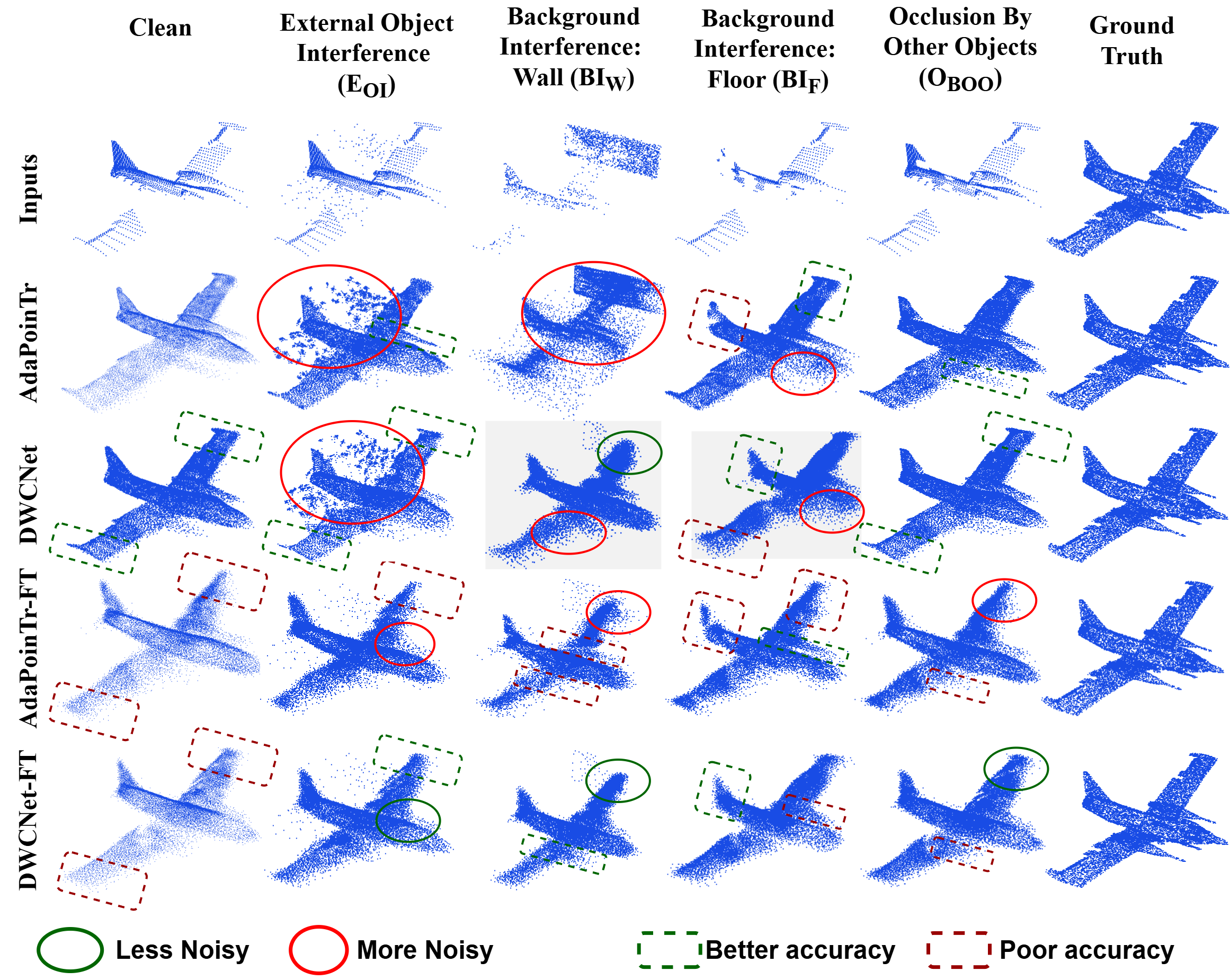}}
\caption{Comparison of AdaPoinTr and DWCNet: Aeroplane Category ($E_{OI}, BI_w , BI_F, O_{BOO}$)} 
\label{01dwca}
\vspace{-0.1in}
\end{figure*}

\begin{figure*}[th!]
\vspace{-0.1in}
\centerline{\includegraphics[width=\linewidth]{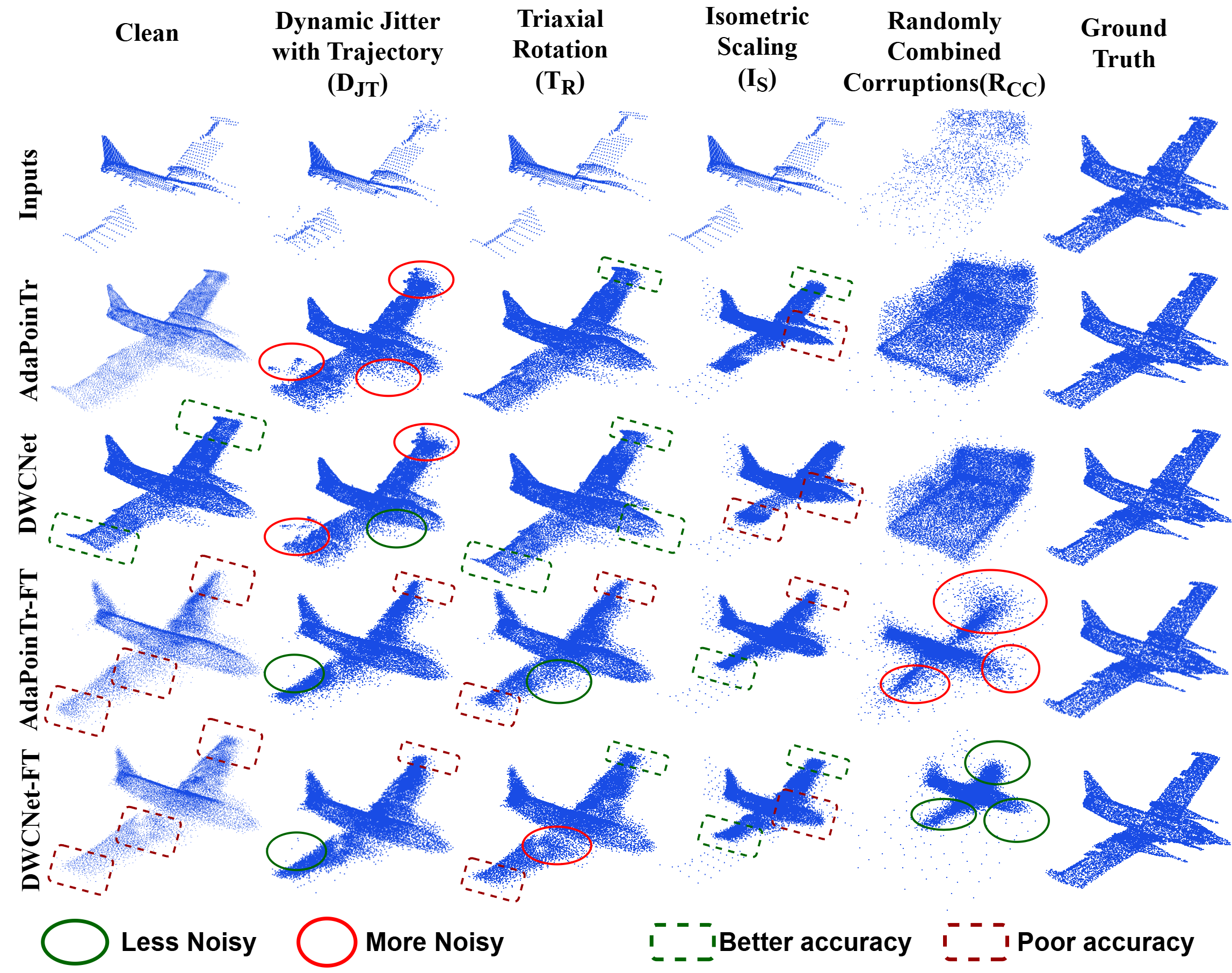}}
\caption{Comparison of AdaPoinTr and DWCNet: Aeroplane Category ($D_{JT}, T_R, I_S, R_{CC}$)} 
\label{01dwcb}
\vspace{-0.1in}
\end{figure*}

\begin{figure*}[th!]
\vspace{-0.1in}
\centerline{\includegraphics[width=\linewidth]{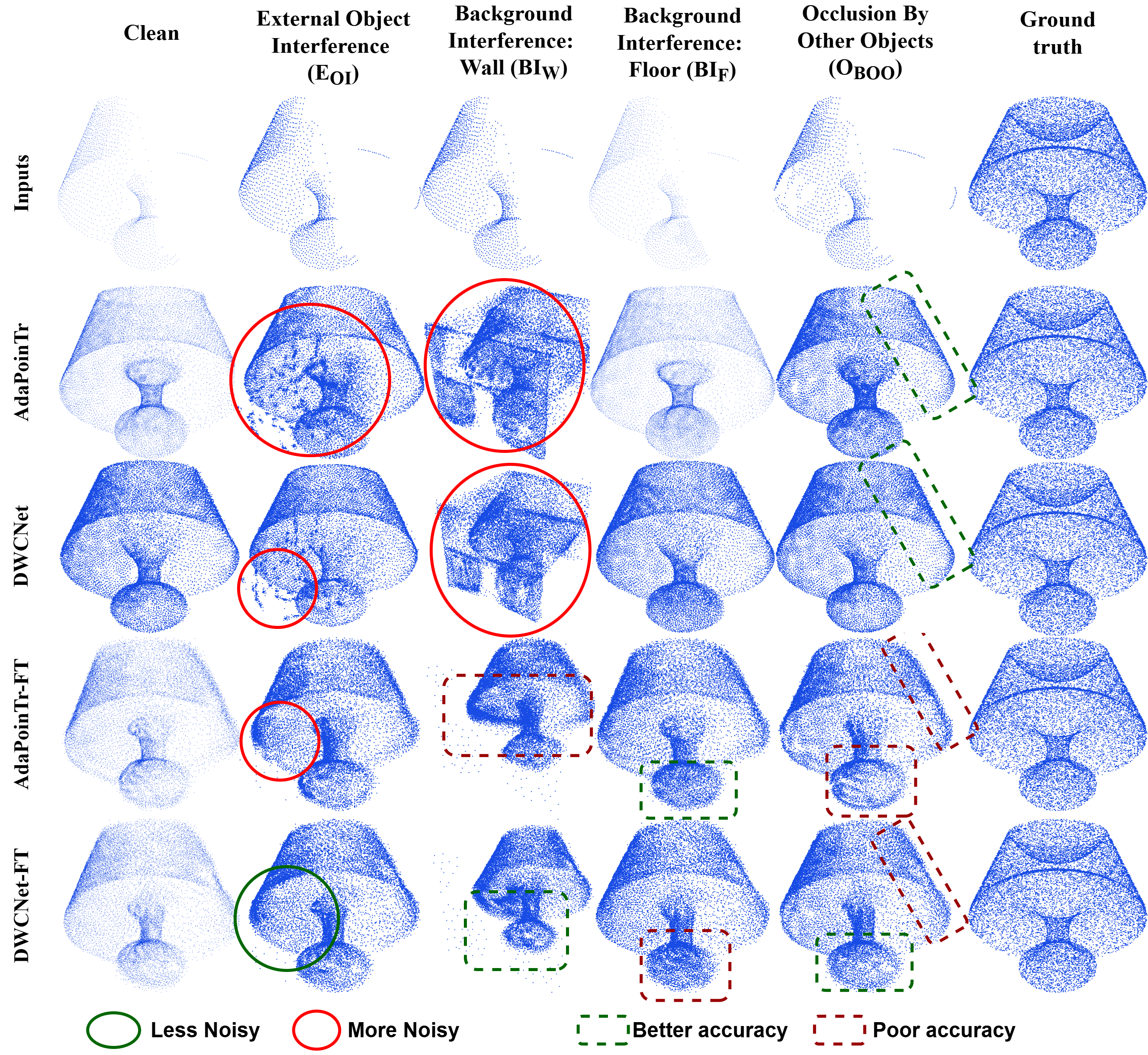}}
\caption{Comparison of AdaPoinTr and DWCNet: Lamp Category ($E_{OI}, BI_w , BI_F, O_{BOO}$)} 
\label{05dwca}
\vspace{-0.1in}
\end{figure*}

\begin{figure*}[th!]
\vspace{-0.1in}
\centerline{\includegraphics[width=\linewidth]{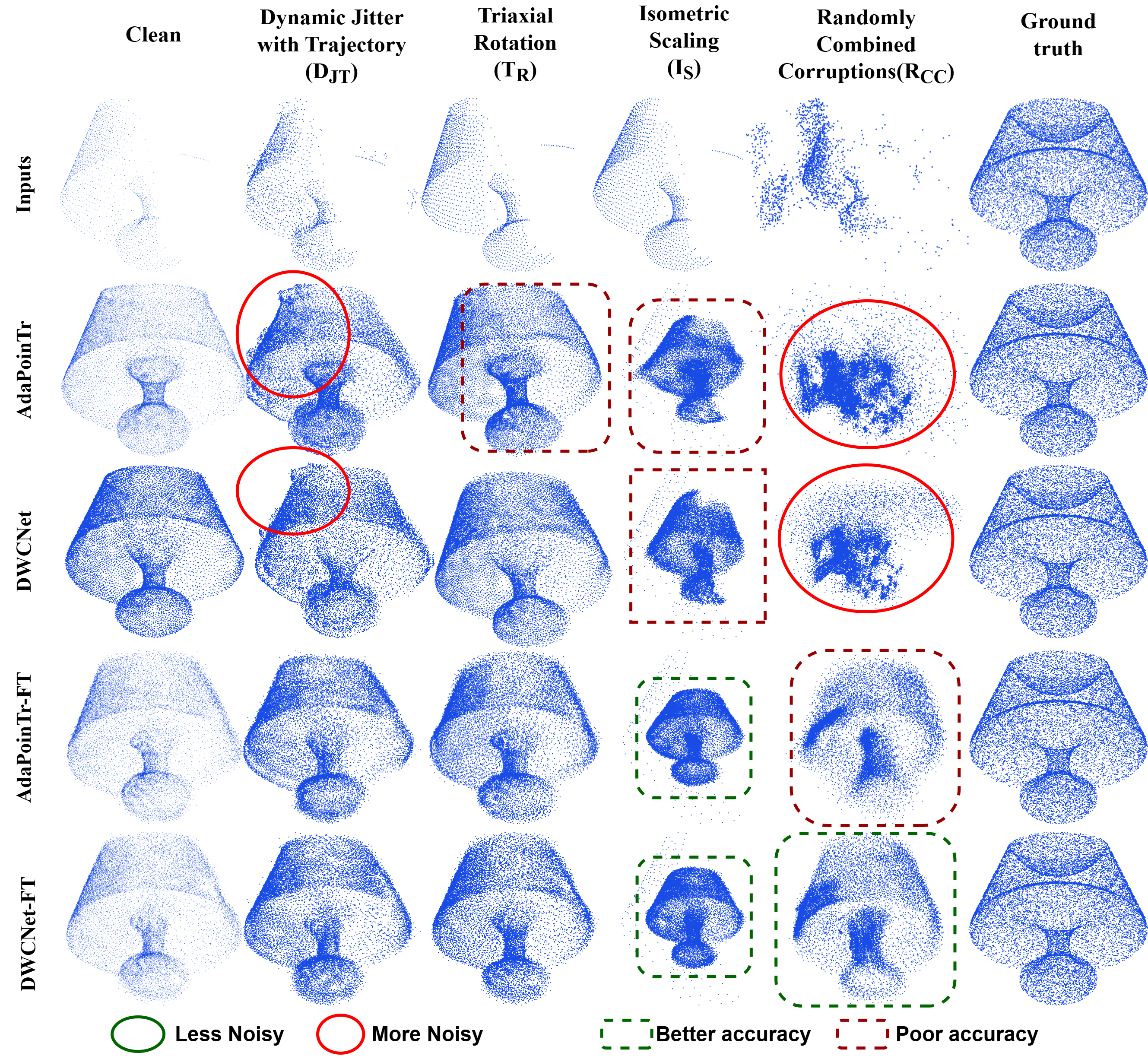}}
\caption{Comparison of AdaPoinTr and DWCNet: Lamp Category ($D_{JT}, T_R, I_S, R_{CC}$)} 
\label{05dwcb}
\vspace{-0.1in}
\end{figure*}

\begin{figure*}[th!]
\vspace{-0.1in}
\centerline{\includegraphics[width=\linewidth]{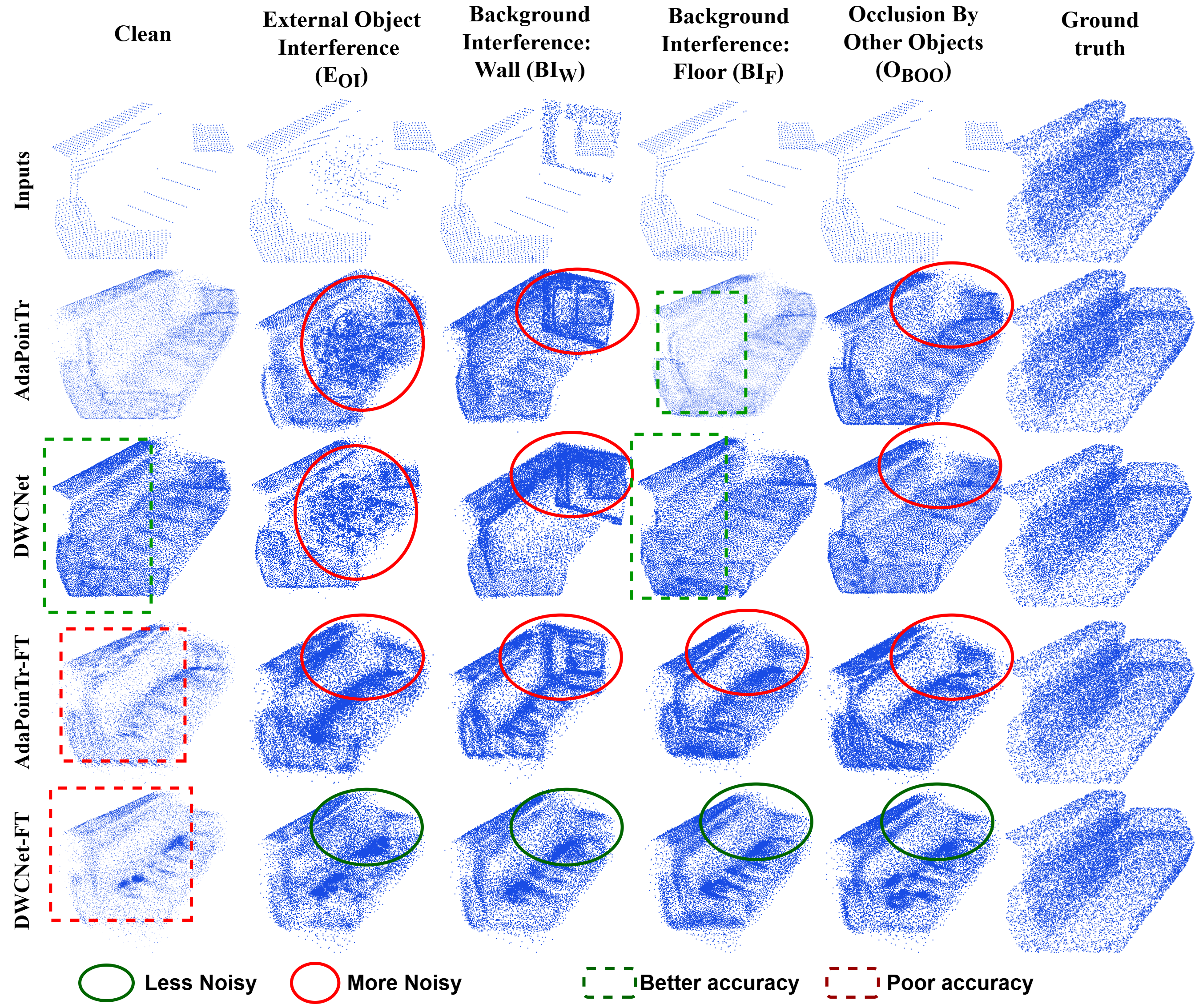}}
\caption{Comparison of AdaPoinTr and DWCNet: Sofa Category ($E_{OI}, BI_w , BI_F, O_{BOO}$)} 
\label{06dwca}
\vspace{-0.1in}
\end{figure*}

\begin{figure*}[th!]
\vspace{-0.1in}
\centerline{\includegraphics[width=\linewidth]{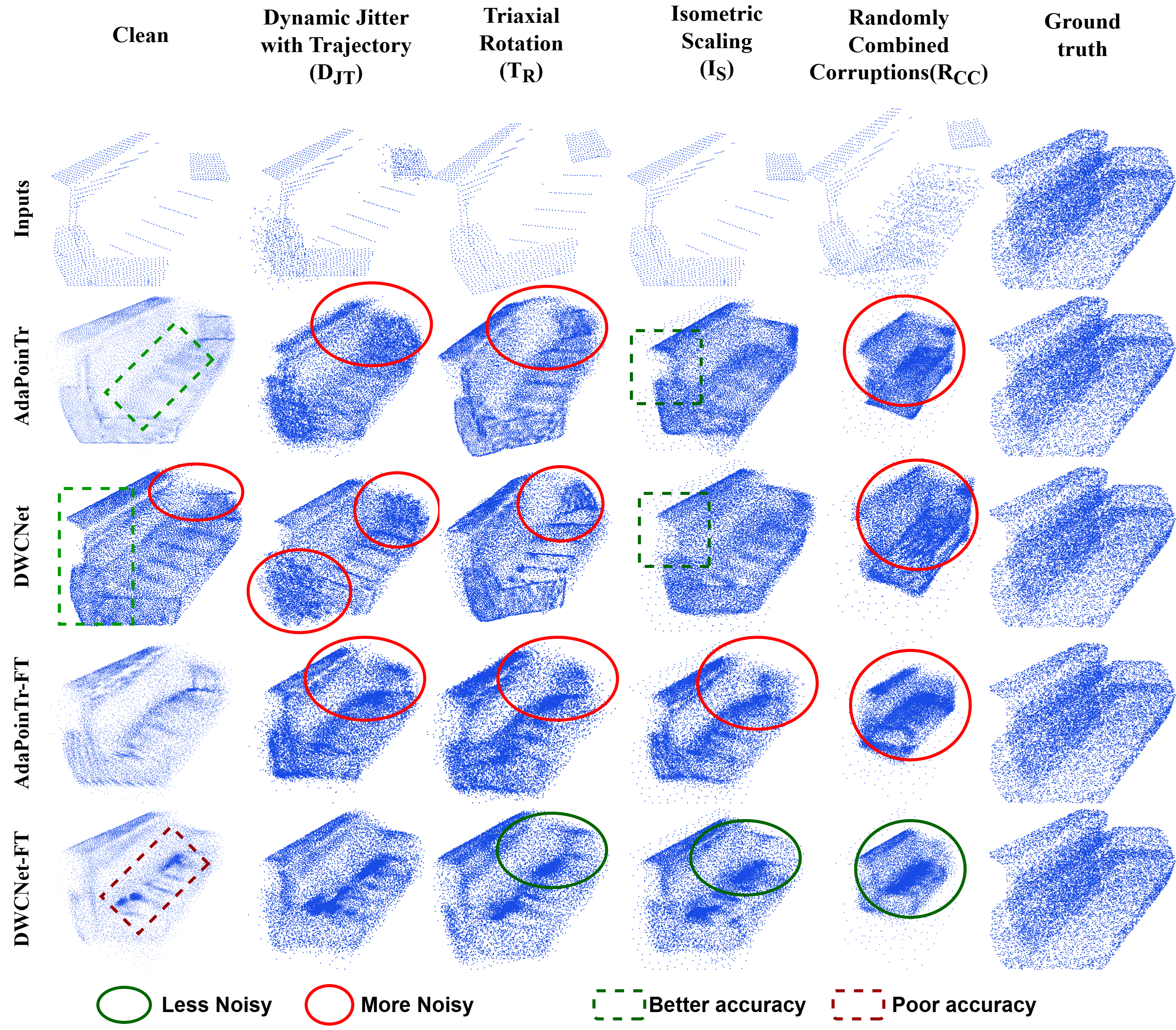}}
\caption{Comparison of AdaPoinTr and DWCNet: Sofa Category ($D_{JT}, T_R, I_S, R_{CC}$)} 
\label{06dwcb}
\vspace{-0.1in}
\end{figure*}

\begin{figure*}[th!]
\vspace{-0.1in}
\centerline{\includegraphics[width=\linewidth]{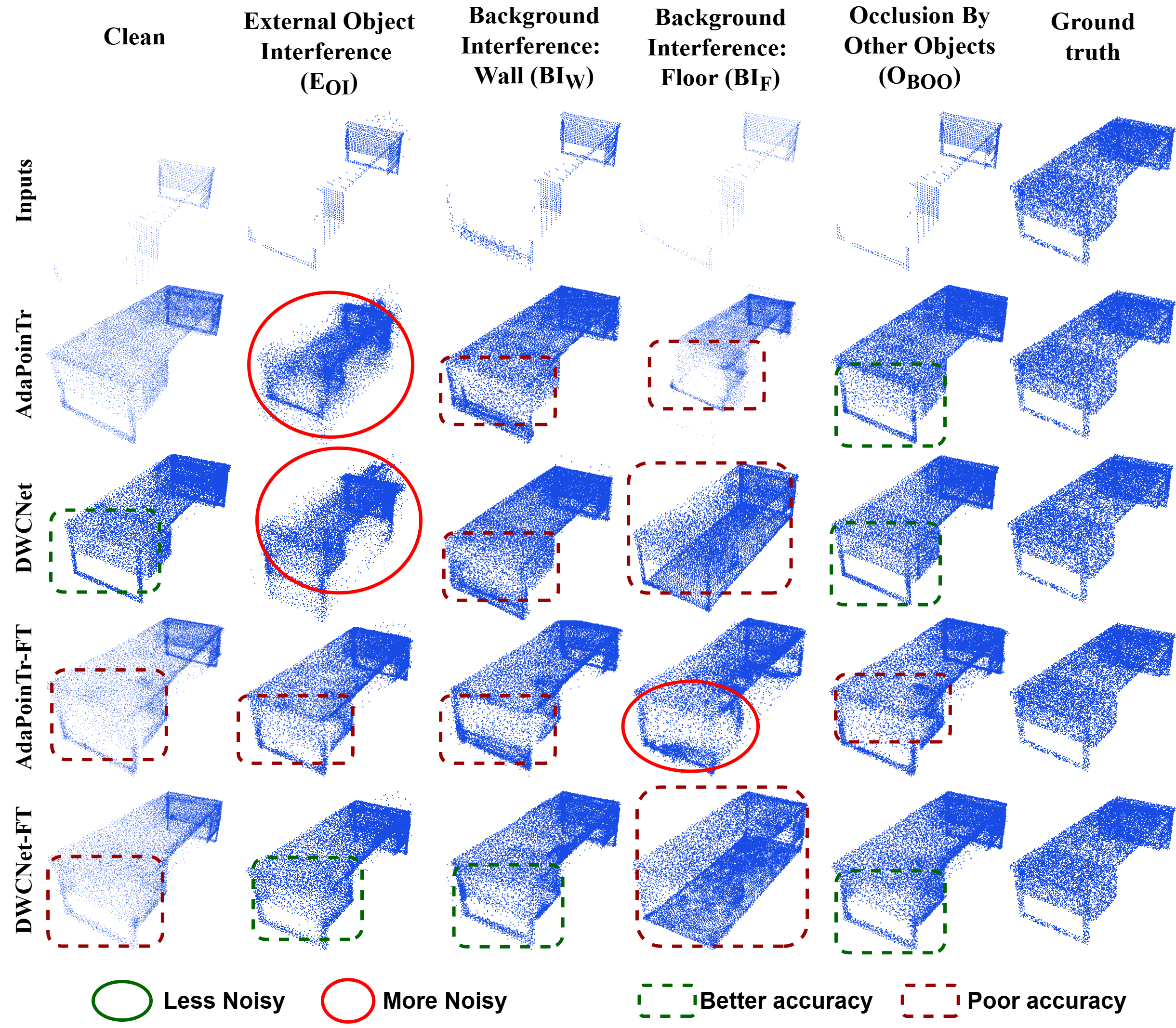}}
\caption{Comparison of AdaPoinTr and DWCNet: Table Category ($E_{OI}, BI_w , BI_F, O_{BOO}$)} 
\label{08dwca}
\vspace{-0.1in}
\end{figure*}

\begin{figure*}[th!]
\vspace{-0.1in}
\centerline{\includegraphics[width=\linewidth]{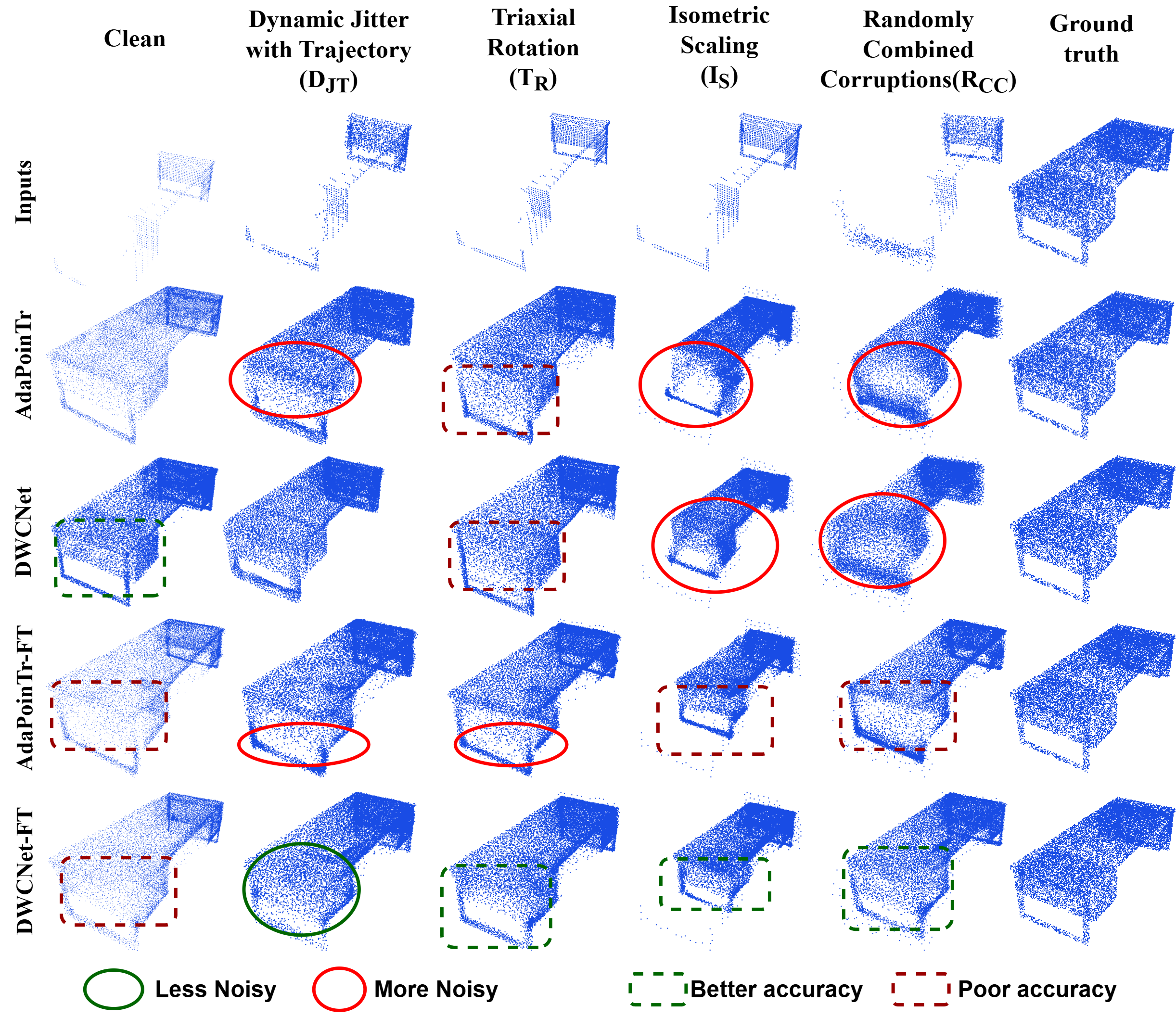}}
\caption{Comparison of AdaPoinTr and DWCNet: Table Category ($D_{JT}, T_R, I_S, R_{CC}$)} 
\label{08dwcb}
\vspace{-0.1in}
\end{figure*}

More qualitative results comparing all networks on all corruptions before and after fine-tuning are shown in figures \ref{01dwcall}, \ref{02dwcall}, \ref{03dwcall} \ref{04dwcall}. More qualitative results comparing AdaPoinTr to DWCNet can be found in figures \ref{02dwca},\ref{02dwcb}, \ref{03dwca},\ref{03dwcb},\ref{04dwca},\ref{04dwcb}, \ref{01dwca}, \ref{01dwcb},\ref{05dwca}, \ref{05dwcb}, \ref{06dwca}, \ref{06dwcb}, \ref{08dwca} and \ref{08dwcb}.

\begin{figure*}[th!]
\vspace{-0.1in}
\centerline{\includegraphics[width=\linewidth]{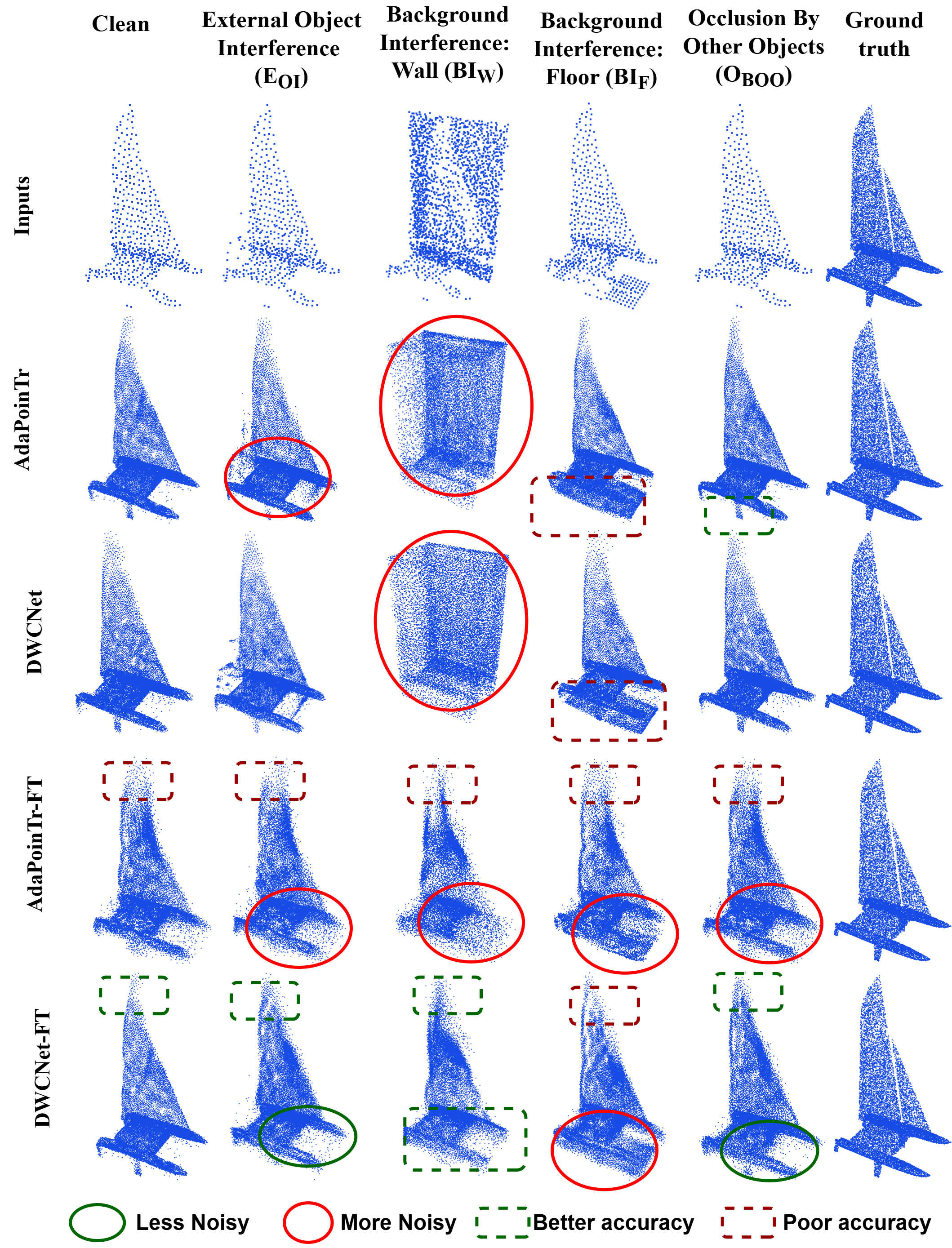}}
\caption{Comparison of AdaPoinTr and DWCNet: Boat Category ($E_{OI}, BI_w , BI_F, O_{BOO}$)} 
\label{15dwca}
\vspace{-0.1in}
\end{figure*}

\begin{figure*}[th!]
\vspace{-0.1in}
\centerline{\includegraphics[width=\linewidth]{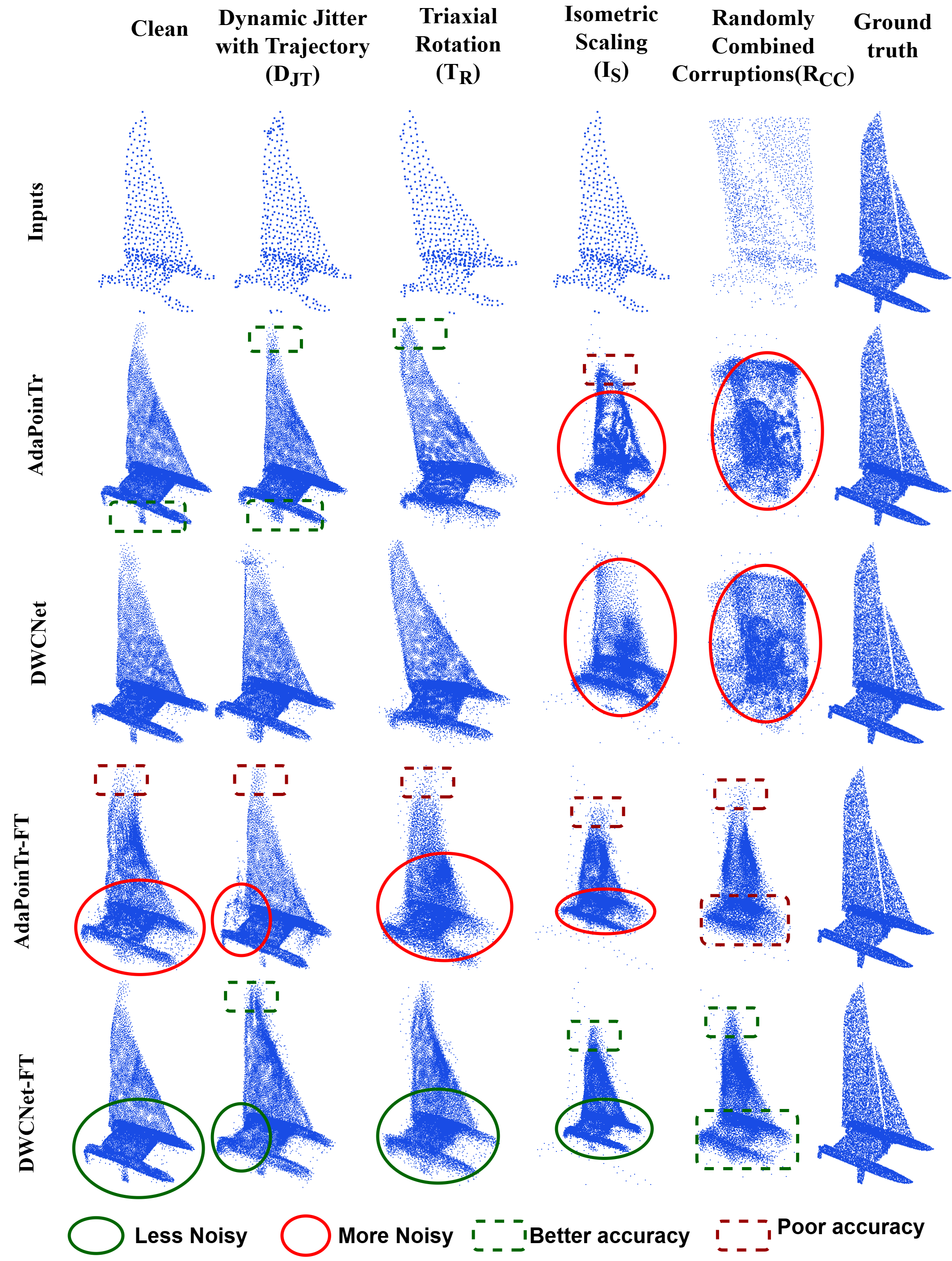}}
\caption{Comparison of AdaPoinTr and DWCNet: Boat Category ($D_{JT}, T_R, I_S, R_{CC}$)} 
\label{15dwcb}
\vspace{-0.1in}
\end{figure*}

\begin{table*}
\caption{Number of Objects, Complete Point Clouds (CPC), and Partial Point Clouds(PPC) in the CPCCD dataset. PC/Corruption refers to number of point clouds per corruption, and No of PC in CPCCD refers to total number of point clouds in CPCCD dataset.}
\label{cpccd_stats}
 \begin{scriptsize}
    \centering
    \setlength{\tabcolsep}{1.4pt}
    {\small
\resizebox{\linewidth}{!}{%
    {\begin{tabular}{c c c c c c c c c c c c}
        \toprule 
         Dataset & Airplane & Cabinet & Car & Chair & Lamp & Sofa & Table & Boat & TNo(objects) & TNo(PPC) & TNo(CPC) \\
        \hline
        Train & 3795 & 1322 & 5677 & 5750 & 2068 & 2923 & 5750 & 1689 & 28974 & 231792 & 28974 \\
        Val & 100 & 100 & 100 & 100 & 100 & 100 & 100 & 100 & 800 & 800 & 800 \\
        Test & 150 & 150 & 150 & 150 & 150 & 150 & 150 & 150 & 1200 & 1200 & 1200 \\
        PC/Corruptions & 4045 & 1572 & 5927 & 6000 & 2318 & 3173 & 6000 & 1939 & 30974 & 233792 & 30974 \\
        \hline
        (No of PC in CPCCD) & 32360 & 12576 & 47416 & 48000 & 18544 & 25384 & 48000 & 15512 & 247792 & 1870336 & 247792 \\
        \hline
        \end{tabular}    
    }}}
    \end{scriptsize}
\end{table*}

\section{CPCCD Dataset}
\label{appendix1}

In this section, we discuss the contents of the CPCCD dataset. The number of objects and point clouds in the datasets are shown in Table~\ref{cpccd_stats}. 

\section{Miscellaneous: Qualitative results for GRNet}
\label{appendix4}
\begin{figure*}[th!]
\vspace{-0.1in}
\centerline{\includegraphics[width=0.9\linewidth]{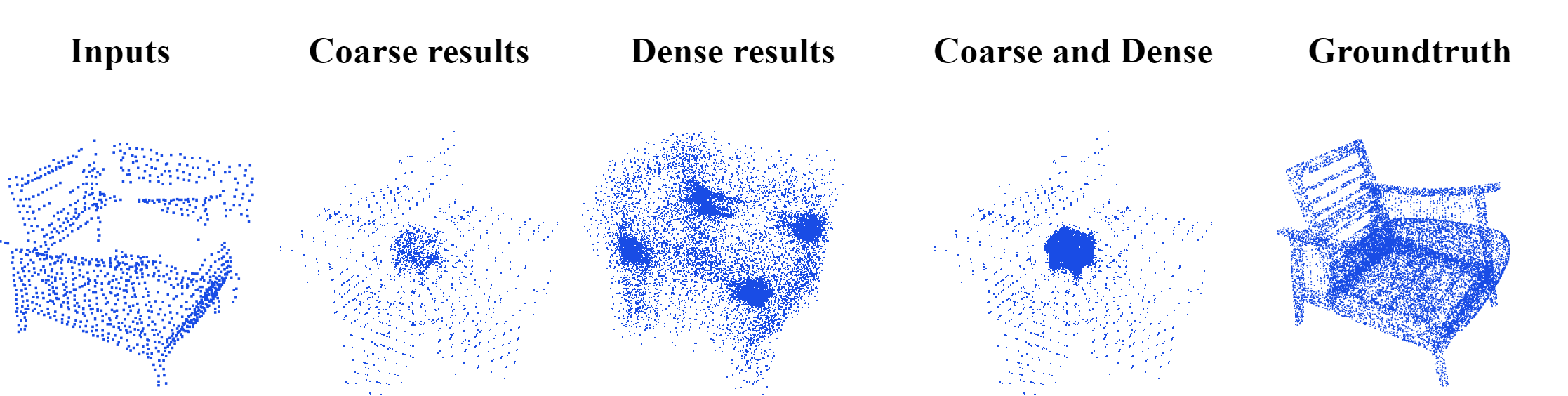}}
\caption{Comparison of the dense and coarse completion results of GRNet} 
\label{01grnet}
\vspace{-0.1in}
\end{figure*}

We provide some qualitative results that provide insight for the performance of GRNet in Figure~\ref{01grnet}. Although the dense results were tolerable, the alignment of the coarse and dense results were mismatched. 

\end{document}